\documentclass[11pt]{svproc}

\usepackage{url}
\usepackage{color}
\usepackage{amsmath}
\usepackage{amssymb}
\usepackage{mathtools}
\usepackage{algorithm}
\usepackage{algpseudocode}
\usepackage{hyperref}
\usepackage{subcaption}
\usepackage{float}

\usepackage{fullpage}

\newcommand{\algName}{ALPaCA}
\newcommand{\algNameLong}{Adaptive Learning for Probabilistic Connectionist Architectures}

\DeclareMathOperator{\Tr}{tr}

\newcommand{\Dj}{D^{(j)}}
\newcommand{\DB}{D^*}
\newcommand{\K}{K} 
\newcommand{\Kb}{\bar{K}} 
\newcommand{\Q}{Q} 
\newcommand{\w}{w} 
\newcommand{\KL}{D_\textrm{KL}}
\newcommand{\testt}{{\theta^*}}
\newcommand{\E}{\mathbb{E}}
\newcommand{\phidim}{{n_\phi}}
\newcommand{\xdim}{{n_x}}
\newcommand{\ydim}{{n_y}}

\begin{document}
\mainmatter              
\title{Meta-Learning Priors for Efficient Online Bayesian Regression}
\titlerunning{Meta-Learning Priors for Bayesian Regression}  
%
\author{James Harrison, Apoorva Sharma, Marco Pavone}

\authorrunning{Harrison, Sharma, Pavone} 
%
\tocauthor{James Harrison, Apoorva Sharma, Marco Pavone}
\institute{Stanford University\\
\email{\{jharrison,apoorva,pavone\}@stanford.edu}}
\maketitle              

\begin{abstract}
Gaussian Process (GP) regression has seen widespread use in robotics due to its generality, simplicity of use, and the utility of Bayesian predictions. The predominant implementation of GP regression is a nonparameteric kernel-based approach, as it enables fitting of arbitrary nonlinear functions. However, this approach suffers from two main drawbacks: (1) it is computationally inefficient, as computation scales poorly with the number of samples; and (2) it can be data inefficient, as encoding prior knowledge that can aid the model through the choice of kernel and associated hyperparameters is often challenging and unintuitive. In this work, we propose \algName{}, an algorithm for efficient Bayesian regression which addresses these issues. \algName{} uses a dataset of sample functions to learn a domain-specific, finite-dimensional feature encoding, as well as a prior over the associated weights, such that Bayesian linear regression in this feature space yields accurate online predictions of the posterior predictive density. These features are neural networks, which are trained via a meta-learning (or ``learning-to-learn'') approach. \algName{} extracts all prior information directly from the dataset, rather than restricting prior information to the choice of kernel hyperparameters. Furthermore, by operating in the weight space, it substantially reduces sample complexity. We investigate the performance of \algName{} on two simple regression problems, two simulated robotic systems, and on a lane-change driving task performed by humans. We find our approach outperforms kernel-based GP regression, as well as state of the art meta-learning approaches, thereby providing a promising plug-in tool for many regression tasks in robotics where scalability and data-efficiency are important.


\end{abstract}

\section{Introduction}

Gaussian Process (GP) regression has become a workhorse in robotics due to its generality, simplicity of use, and the utility of Bayesian predictions. As such, it has been used in robotic applications as diverse as model identification for control and reinforcement learning \cite{deisenroth2011pilco,berkenkamp2017safe}, contact modeling \cite{bauza2017probabilistic}, terrain and environment modeling \cite{vasudevan2009gaussian,o2012gaussian}, prediction of human behavior \cite{wang2008gaussian,urtasun20063d}, and trajectory optimization \cite{mukadam2016gaussian}. Moreover, it has been used effectively for essential tools in robotics such as Bayesian filtering \cite{ko2009gp}, SLAM \cite{ferris2007wifi}, and Bayesian optimization \cite{snoek2012practical}.

While GP regression is a useful tool, standard approaches have several well-known limitations. GP regression is predominantly performed via kernel-based methods \cite{rasmussen2004gaussian}, but incorporating physical priors into these models is difficult, and done implicitly through the choice of kernel function. For example, in modeling a physical system with a GP equipped with the widely-used squared exponential (SE) kernel, the system designer may vary the prior variance, and the length scale of the kernel, which controls smoothness \cite{rasmussen2004gaussian}. These choices of priors are neither expressive nor intuitive enough for complex applications in robotics. Moreover, kernel-based GP regression algorithms scale cubically with the amount of data \cite{hensman2013gaussian}. This high sample complexity prevents application beyond relatively small problems \cite{smola2001sparse}.

Several works have attempted to incorporate prior knowledge into GP models and to remove the limitations imposed by kernel methods. Methods mapping input data to alternative feature spaces which are then used as inputs to kernel GP models have been shown to allow incorporation of prior knowledge, and reduce the complexity by projecting to a lower dimensional space \cite{mackay1998introduction}.
For example, Hinton and Salakhutdinov \cite{hinton2008using} use a Deep Belief Network to learn a feature representation in an unsupervised fashion, but this separates the process of representation learning and regression, which may be suboptimal.
Manifold GPs \cite{calandra2016manifold} jointly learn a neural network transformation of the data such that GP methods applied to this feature space achieve good predictions. 
However, this approach relies on the system designer having access to data generated from the exact system being modeled, and it is unclear how well this approach will perform on a  regression problem that differs from that used to generate the training dataset. Good performance on a general class of tasks (as opposed to one specific task within this family) is necessary for effective autonomous operation in unstructured environments. 
Moreover, these approaches still suffer from the poor sample complexity associated with kernel methods. 
A more extensive discussion of methods for incorporating prior knowledge is presented in Section \ref{sec:rw}.

A variety of sparse and approximate methods have been proposed to reduce the sample complexity of kernel methods, but these at best scale quadratically \cite{snelson2006sparse,np2018}. 
A popular approach to avoiding the complexity of kernel methods is to instead turn to random features for approximating kernel functions, and solving the linear regression problem using these features instead \cite{rahimi2008random}. The approach taken in this work is similar, but we learn features that are well-suited to representing posterior densities, based on available prior information.

\paragraph{Contributions.}

In this work, we propose \algName{} (\algNameLong{}), an algorithm for incorporating prior information in to GP regression that increases computational and sample efficiency, and provides Bayesian posteriors reflecting confidence in predictions.
We propose to learn a neural network model (i.e. connectionist architecture) offline that is trained to be able to rapidly incorporate online information to predict posterior distributions.
Online adaptation is achieved by performing Bayesian linear regression on the last layer of the network, and thus our approach is capable of providing confidence intervals for predictions. 
Fundamentally, our network learns priors over the last layer weights, as well as network weights, such that it can optimally predict the posterior over the label, conditioned on the online samples observed so far. 
This is performed by leveraging the ``learning-to-learn'' approach of meta-learning \cite{finn2017model}, where model training is done to minimize divergence between the predicted and true posterior, for any amount of observed online context data.
By performing regression on meta-learned basis functions, \algName{} achieves complexity during inference that is \textit{linear} in the amount of data. 
As such, \algName{} offers a general, plug-in alternative to kernel GP regression which can improve scalability and online data-efficiency in a wide variety of robotics applications.


\paragraph{Organization.}
In Section 2, we provide a formal problem statement. We present background information on Bayesian linear regression and GP regression in Section 3. In Section 4 we introduce the \algName{} algorithm for generic supervised learning tasks. We contrast our approach to the literature on neural network regression models, current meta-learning approaches, as well as popular GP models in Section 5. Finally in Section 6, we demonstrate \algName{} on simple problems that are standard benchmarks in meta-learning \cite{finn2017model}, as well as high-dimensional robotic systems involving contact, and prediction of human decision making in a driving task\footnote{The code for all of our experiments is available at \url{https://github.com/StanfordASL/ALPaCA}}. \algName{} is shown to outperform kernel GP regression, as well as meta-learning approaches such as MAML \cite{finn2017model}.

\section{Problem Formulation}
In this work, we are focused on the problem of Bayesian function regression. In this setting, we start with a prior over possible functions, observe samples from an unknown test function, and use these samples to compute a posterior predictive density over the unknown function.

Formally, let $f: \mathbb{R}^\xdim \to \mathbb{R}^\ydim$ denote an arbitrary, possibly nonlinear and nonsmooth function with latent parameters $\theta$. 
We consider a setting where $f$ is fixed and captures problem specific structure, while $\theta$ is uncertain and explains intra-domain variation.
For example, in the context of prediction of human driving behavior, uncertainty in the state evolution stems from uncertainty in the driver's intent, while the physics governing the car's dynamics are known.
We assume the samples of $f$ observed online may be corrupted with additive Gaussian noise. That is, we observe a sequence of samples $(x,y)$ such that
\begin{equation}
    p(y\mid x, \theta) = \mathcal{N}(f(x; \theta), \Sigma_\epsilon),
    \label{eq:dynamics}
\end{equation}
where $\mathcal{N}(\cdot,\cdot)$ denotes the multivariate Gaussian distribution, and $\Sigma_\epsilon$ is the noise covariance, and $\theta$ is unknown. 

Suppose we observe a collection of $\tau$ sample points $\DB_\tau = \{(x_t,y_t)\}_{t=1}^{\tau}$ coming from an unknown function with latent parameter $\testt$.
Given a prior over model parameters $p(\theta)$, and this context data $\DB_\tau$, the posterior predictive density over $y$ given $x$ can be computed as
\begin{equation}
    p(y\mid x, \DB_\tau) = \int p(y \mid x, \theta) p(\theta \mid \DB_\tau)\,d\theta.
\end{equation}
Computing this posterior predictive density, however, requires having analytic expressions for $f(x;\theta)$ and $p(\theta)$, and even then, computing the posterior $p(\theta \mid \DB_\tau)$ exactly may remain intractable. 
Instead, we propose using a surrogate model for which computing the posterior predictive density is analytically tractable, and optimizing this surrogate model \textit{offline} such that its posterior preditictive density is close to the true posterior predictive density for all likely settings of $\testt$ and the resulting observations $\DB_\tau$. 


Let $q_\xi(y \mid x, \DB_\tau)$ represent the posterior predictive density under our surrogate model, parameterized by $\xi$. Let $\hat{y}$ represent a sample from this density $q$. 
We consider a scenario where data comes in as a stream: that is, at each timestep, the agent is presented with an input $x_t$, and after estimating the output $\hat{y}_t$, the agent is provided with the true value $y_t$. 
An example of this is a Markovian dynamical system, in which an agent wishes to predict the distribution over the next state, and at the next time step, this state is observed.

In this context, the problem we wish to solve can be written as
\begin{equation}
 \min_{\xi}~~ \KL\left(p(y \mid x_{t+1}, \DB_{t})\, \| \,q_\xi(y \mid x_{t+1}, \DB_{t})\right),
\label{eq:div}
\end{equation} 
where $\KL(\cdot \| \cdot)$ denotes the Kullback-Leibler (KL) divergence. Note that this objective is minimizing the conditional KL divergence. As such, the objective is to minimize this divergence in expectation over $t, x_{t+1}, D^*_{t}$, and $\testt$. 
Directly optimizing this objective requires access to $p(\theta)$ and $p(\DB_t \mid \testt)$.
Instead of assuming access to these distributions directly, we only require a dataset of samples drawn from different settings of latent parameter $\theta$. Specifically, we will write $\Dj_\tau \coloneqq \{(x_t,y_t)\}_{t=1}^{\tau}$ to represent a dataset of $\tau$ samples generated via sampling (iid) $\theta_j \sim p(\theta)$, $x_t \sim p(x)$ and $y_t \sim p(y\mid x_t, \theta_j)$, that is, a set of samples from one function instantiation. Let $\mathcal{D} \coloneqq \{\Dj_\tau\}_{j=1}^{M}$ represent a collection of these datasets. Note that by sampling $t\sim p(t)$, sampling $\Dj_\tau$ uniformly from $\mathcal{D}$, and the truncating $\Dj_\tau$ to $t<\tau$ elements, we can approximate the expectations in (\ref{eq:div}) through a Monte Carlo scheme.

\section{Bayesian Regression}

In this work, we utilize Bayesian multivariate linear regression as a tool for computing $q(y\mid x, \DB_\tau)$, as it allows for analytic computation of the posterior. In this section, we offer a brief review of the technique.
For a more in-depth treatment, we refer the reader to \cite{rasmussen2004gaussian}, \cite{minka2000bayesian}, or \cite{murphy2012machine}. Let $x_t \in \mathbb{R}^\xdim$ and $y_t  \in \mathbb{R}^\ydim$ denote the independent and dependent variables respectively. We will consider regression over basis functions $\phi: \mathbb{R}^\xdim \to \mathbb{R}^\phidim$. Then, $y_t^T = \phi^T(x_t) \K + \epsilon_t$, where $\K \in \mathbb{R}^{\phidim \times \ydim}$ is a coefficient matrix and $\epsilon \sim \mathcal{N}(0,\Sigma_\epsilon)$, wherein $\mathcal{N}(\cdot,\cdot)$ denotes the multivariate Gaussian distribution and $\Sigma_\epsilon$ denotes the noise variance. We will assume $\Sigma_\epsilon$ is known throughout this work. The case is which this assumption is relaxed is discussed in the appendix. Let 
$Y^T = [y_{1}, \ldots, y_{\tau}]$,
$\Phi^T = [\phi(x_1), \ldots, \phi(x_\tau)]$, where $\tau$ is the number of data points.
Stacking samples $t = 1, \ldots, \tau$, we may equivalently write $Y = \Phi K + E$. 
Given this formulation, the conditional probability of the data $Y$ is
\begin{equation}
    p(Y \mid \Phi, \K, \Sigma_\epsilon) \propto |\Sigma_\epsilon|^{-\ydim/2} \exp\left(-\frac{1}{2} \Tr \left(( Y - \Phi \K) \Sigma_\epsilon^{-1} (Y - \Phi \K)^T \right)\right),
    \label{eq:likelihood}
\end{equation}
where $\Tr(\cdot)$ denotes the trace operation. A natural conjugate prior for $\K$ is
$    p(\K) \sim \mathcal{MN}(\Kb_0,\Lambda_0^{-1},\Sigma_\epsilon),
$ where $\mathcal{MN}$ denotes the matrix normal distribution, and $\Lambda_0$ is a precision matrix (the inverse of the variance). The matrix normal distribution is equivalent to the multivariate normal distribution with mean $\textrm{vec}(\Kb_0)$ and variance $\Sigma_\epsilon \otimes \Lambda_0^{-1}$, where $\textrm{vec}(\Kb_0)$ represents the vectorization of $\Kb_0$ and $\otimes$ denotes the Kronecker product.
The posterior, conditioned on $Y$ and $\Phi$, is then 
$    p(\K \mid Y,\Phi) = \mathcal{MN}(\Kb_\tau,\Lambda_\tau^{-1},\Sigma_\epsilon),
$ where
\begin{align}
    \Lambda_\tau &= \Phi^T \Phi + \Lambda_0  \label{eq:updates1}\\
    \Kb_\tau &= (\Lambda_\tau)^{-1} (\Phi^T Y + \Lambda_0 \Kb_0).
    \label{eq:updates2}
\end{align}
Given the above, the posterior predictive distribution is
\begin{equation}
    p(y_{\tau+1}\mid \phi(x_{\tau+1}), \Phi,Y) = \mathcal{N}(\Kb_\tau^T \phi(x_{\tau+1}), \Sigma_{\tau+1})
    \label{eq:s_like}
\end{equation}
where 
\begin{equation}
    \Sigma_{\tau+1}  = (1 + \phi^T(x_{\tau+1}) \Lambda^{-1}_\tau \phi(x_{\tau+1})) \Sigma_\epsilon
    \label{eq:sk}.
\end{equation}
For a derivation of this term, the reader may refer to \cite{minka2000bayesian}\footnote{While the expressions in this work and \cite{minka2000bayesian} are not obviously equivalent, they can be shown to be the same by applying the Woodbury identity.}. Note that several common forms of linear regression, such as ridge regression, are special cases of this general formulation \cite{murphy2012machine}.


Bayesian linear regression has strong connections to kernel methods for GP regression. With the choice of Gaussian priors discussed here, Bayesian linear regression implicitly defines a GP with mean and kernel functions that are functions of the prior on the weights, $p(K)$, and the basis functions used. 

Note that the terms involving the basis functions $\phi$ are always weighted inner products of $\phi(x)$ and $\phi(x')$ (with the same weighting, which we will denote $\Sigma$). Thus, we define the kernel function $k(x,x') = \phi^T(x) \Sigma \phi(x')$. 
In contrast to Bayesian linear regression, kernel-based GP regression employs the well-known ``kernel trick'' to replace occurrences of inner products with the kernel function \cite{rasmussen2004gaussian}. In this way, it can effectively operate in an infinite dimensional feature space such as that defined by the popular squared exponential kernel. This added model capacity comes at the cost of poor computational complexity as the number of samples in the dataset grows.


\begin{figure}[t]
    \centering
    \includegraphics[width=0.7\columnwidth]{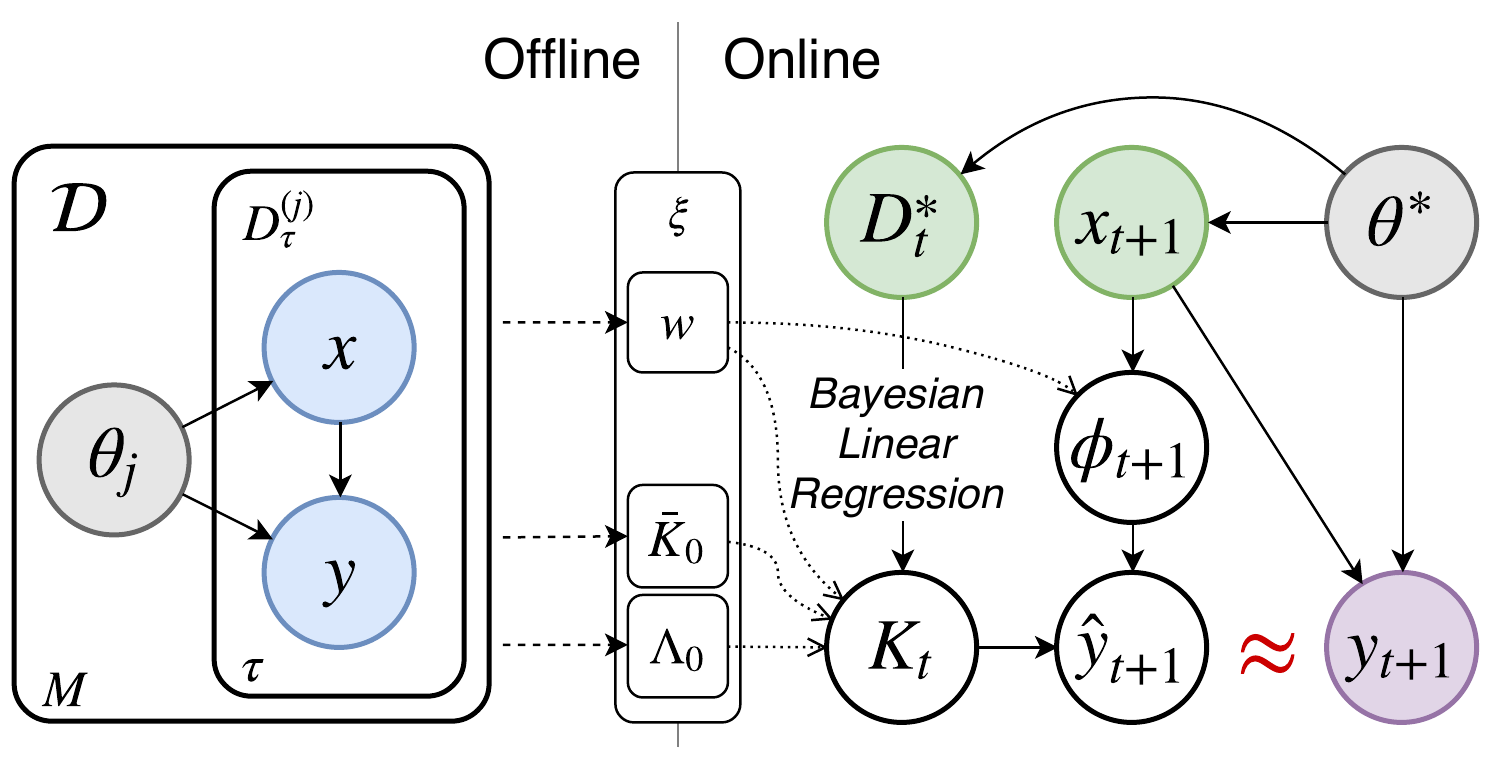}

    \caption{Overview of the \algName{} algorithm. 
    Rather than compute the intractable posterior $p(y_{t+1} \mid x_{t+1}, \DB_t)$ directly, we approximate it through a surrogate model for which computing the posterior $q_\xi(\hat{y}_{t+1} \mid x_{t+1}, \DB_t)$ is analytically tractable via Bayesian linear regression. Offline, we optimize the parameters $\xi$ of this surrogate model on a dataset $\mathcal{D}$ of functions drawn from a prior distribution, aligning the surrogate posterior with the true empirical posterior in the dataset through a meta-learning procedure.
    Blue nodes are observed offline, green are online inputs, and purple the targets. White nodes are computed, and gray nodes are not observed.}
    \label{fig:pgm-model}
\end{figure}

\section{Approach}

In this work, we choose to represent the approximate density $q(y\mid x_{t+1}, \DB_t)$ as a neural network model with a Bayesian last layer, which allows employing Bayesian linear regression to compute $q(y\mid x_{t+1}, \DB_t)$ analytically. Section \ref{sec:algo_prelims} provides details of this model and other preliminaries. In Section \ref{sec:algo_overview}, we outline the full \algName{} algorithm.

\subsection{Algorithmic Preliminaries}
\label{sec:algo_prelims}


\paragraph{Network Model.} Recall $\hat{y}_t$ denotes the random variable distributed according to $q(y \mid x, \DB_\tau)$. In this work we will learn a parametric model for $\hat{y}_t$,
\begin{equation}
\hat{y}_{t} = \K^T \phi(x_t; \w) + \epsilon
\label{eq:model},
\end{equation}
where $\phi$ is a neural network with output dimensionality $n_\phi$ and weights $\w$, $\epsilon \sim \mathcal{N}(0,\Sigma_\epsilon)$, and $\K$ is a matrix which may be thought of as the last layer of the neural network. Learning this model based on a dataset for fixed $\theta$ is relatively simple, and is in line with the majority of the literature on neural network regression models. However, we wish to learn a prior distribution over $\K$ such that given data sampled from (\ref{eq:dynamics}) with the test parameter $\testt$, we can generate accurate Gaussian posteriors. Specifically, we wish to learn the parameters $\Kb_0, \Lambda_0$ of a matrix normal prior $p(K) = \mathcal{MN}(\Kb_0, \Lambda_0^{-1}, \Sigma_\epsilon)$, so that we may apply tools from Bayesian linear regression to analytically compute posterior distributions given online data. 

While a fully Bayesian approach would also aim to estimate the posterior over the weights $\w$, we will assume a delta prior over these weights, and thus they are not updated online. However, note that in comparison to previous work that utilizes networks with Bayesian output layers \cite{snoek2015scalable,bauer}, we will not train the network and learn the prior over $\K$ in separate phases, and instead they are learned jointly. This is an important distinction: as opposed to simply learning a network to produce good predictions in expectation (with no context data), and adding in a Bayesian last layer afterwards, we aim to learn a network that is composed of basis functions capable of representing the posterior distribution given any context data. Given fixed basis functions, learning the prior terms in a data-driven fashion is known as empirical Bayes \cite{morris1983parametric}. In contrast to the empirical Bayes approach, we also learn the basis functions in addition to the matrix normal prior. Thus, in summary, the parameters of the posterior predicitve model $q_\xi$ we wish to optimize are $\xi = (\Kb_0, \Lambda_0, \w)$.
 


 \paragraph{Optimization Objective.}

We will denote the process generating $\DB$, data $x$, and label $y$ as $p(x,y,\DB \mid \testt)$, for online test parameters $\testt$. 
 Note that this may equivalently be written as $p(y \mid x, \testt) p(x, \DB \mid \testt)$, as $y$ is conditionally independent of $\DB$, given $\testt$.
 Making explicit the dependence of the posterior on the prior terms, the inner expectation of (\ref{eq:div}), wherein the expectation over $\testt$ and $t$ is temporarily ignored, may be written
\begin{align*}
    \E_{x,\DB \sim p(x,\DB \mid \testt)}\left[  \mathbb{E}_{y\sim p(y\mid x,\testt)}[ \log p(y \mid x, \DB) - \log q_\xi(y \mid x, \DB) \mid x, \DB ] \right].
\end{align*}
We can then write the above divergence, dropping the terms that do not affect the optimization over the prior terms, as
\begin{align*}
    L(\Kb_0, \Lambda_0, \w, \testt)
    = - \E_{x,y,\DB \sim p(x,y,\DB \mid \testt)} \left[ \log q_\xi(y \mid x, \DB) \right].
\end{align*}
Rewriting this negative log likelihood using (\ref{eq:s_like}), discarding terms not relevant to the optimization problem over the prior terms, and making the time dependency explicit, we obtain
\begin{align*}
     L &(\Kb_0, \Lambda_0, \w, \testt, t) =\\
     &\E_{x_{t+1},y_{t+1},\DB_{t} \sim p(x,y,\DB \mid \testt)} \left[ \log \det (\Sigma_{t+1})
     + (y_{t+1} - \Kb_{t}^T \phi_{t+1})^T \Sigma_{t+1}^{-1} (y_{t+1} - \Kb_{t}^T \phi_{t+1}) \right].
\end{align*}
Thus, (\ref{eq:div}) may be written as
\begin{equation}
     \ell (\Kb_0, \Lambda_0, \w) = \E_{t \sim p(t), \testt \sim p(\theta)} \left[ L(\Kb_0, \Lambda_0, \w, \testt, t) \right]
     \label{eq:loss}
\end{equation}
where  $\Lambda_t$, $\Kb_t$, and $\Sigma_t$ are as in (\ref{eq:updates1}), (\ref{eq:updates2}), and (\ref{eq:sk}). These terms are functions of the prior terms, $\Kb_0, \Lambda_0$, and $\w$, as well as $\DB_{t}$. Here, we write $\phi_{t}$ to denote $\phi(x_{t})$. 
This equivalence between minimizing KL divergence and maximizing log likelihood is well-known \cite{murphy2012machine}, but we emphasize it to improve clarity regarding the expectations and role of the prior terms in the likelihood.






\subsection{\algName{} Overview}
\label{sec:algo_overview}

\begin{algorithm}[t]
\caption{\label{alg:alpaca} \algName{}: Offline}
\centering
\begin{algorithmic}[1]
\Require training data $\mathcal{D}$, noise variance $\Sigma_\epsilon$
    \State Randomly initialize $\Kb_0, \Lambda_0, \w$ 
    \While{not converged}
        \For{$j=1$ \textbf{to} $J$}
            \State Sample $t_j$ from uniform distribution over $\{1, \ldots, \tau\}$
            \State Sample dataset $D^{(j)}$ uniformly from $\mathcal{D}$
            \State Compute $\Kb_{t_{j}}, \Lambda_{t_{j}}, \Sigma_{t_{j}+1}$ via (\ref{eq:opt_prob}), using dataset $D^{(j)}_{t_j+1}\setminus\{(x_{t_j+1},y_{t_j+1})\}$
        \EndFor
        \State Update $\K_0, \Lambda_0, \w$ via gradient step on the optimization problem (\ref{eq:opt_prob})
    \EndWhile
    \State \Return $\Kb_0, \Lambda_0, \w$
  \end{algorithmic}
\end{algorithm}

\paragraph{Offline Training.}

Given (\ref{eq:loss}), we are now able to write the full optimization problem we wish to solve. We sample $J$ datasets $D_\tau$ of $\tau$ samples each uniformly from $\mathcal{D}$. Let $j$ denote the index of these datasets. For each of these datasets, we sample a horizon $t_j \sim p(t)$ and clip each dataset to this horizon to obtain $\{D^{(j)}_{t_j+1}\}_{j=1}^J$. 
Because $\mathcal{D}$ contains datasets with $\theta\sim p(\theta)$, evaluating the loss on this minibatch approximates the expectations in (\ref{eq:loss}). We use the superscript $(j)$ to denote the terms within each dataset $D^{(j)}_{t_j+1}$.
Note that 

\begin{align*}
    \log \det (\Sigma_{t_j+1}) &= \log \det((1 + \phi^T(x_{t_j+1}) \Lambda^{-1}_{t_j} \phi(x_{{t_j+1}})) \Sigma_\epsilon)\\ 
    &= \ydim \log(1 + \phi^T(x_{{t_j+1}}) \Lambda^{-1}_{t_j} \phi(x_{{t_j+1}})) + \log \det \Sigma_\epsilon.
\end{align*}
Since it does not depend on the optimization variables, $\log \det \Sigma_\epsilon$ may be ignored and we may write the Monte Carlo estimate of (\ref{eq:loss}) as
\begin{equation}
\begin{aligned}
    \hat{\ell}(\Kb_0, \Lambda_0, \w) \coloneqq
    \frac{1}{J}\sum_{j=1}^J \biggl(& \ydim \log \left(1 + \phi^{(j),T}_{t_j+1} \Lambda^{(j),-1}_{t_j} \phi^{(j)}_{t_j+1}\right) \\
     &+ \left(y^{(j)}_{t_j+1} - \Kb^{(j),T}_{{t_j}} \phi^{(j)}_{{t_j}+1}\right)^{T} \Sigma_{t_j+1}^{(j),-1} \left(y^{(j)}_{t_j+1} - \Kb_{{t_j}}^{(j),T} \phi^{(j)}_{{t_j}+1}\right) \biggr).
\end{aligned}
\end{equation}
Thus, our problem formulation is 
\begin{equation}
\begin{aligned}
\label{eq:opt_prob}
& \underset{\Kb_0,\Lambda_0,\w}{\min}
& & \hat{\ell}(\Kb_0, \Lambda_0, \w) \\
& \text{\,\,\, s.t.}
& & \Sigma^{(j)}_{{t_j}+1}  = (1 + \phi^{T}(x^{(j)}_{{t_j}+1}) \Lambda^{(j),-1}_{t_j} \phi(x^{(j)}_{{t_j}+1})) \Sigma_\epsilon, \,\, j = 1, \ldots, J \\
&&& \Lambda^{(j)}_{t_j} = \Phi^{(j),T}_{t_j} \Phi^{(j)}_{t_j} + \Lambda_0, \,\, j = 1, \ldots, J \\
&&& \Kb^{(j)}_{t_j} = (\Lambda^{(j)}_{t_j})^{-1} (\Phi_{t_j}^{(j),T} Y_{t_j}^{(j)} + \Lambda_0 \Kb_0), \,\, j = 1, \ldots, J\\ 
&&& \Lambda_0 \succeq 0
\end{aligned}
\end{equation}
where $\Phi^{(j)}_{t_j}$ and $Y^{(j)}_{t_j}$ contain data from dataset $j$, from timesteps $1, \ldots, t_j $.
The full algorithm is detailed in Algorithm \ref{alg:alpaca}. The semidefinite constraint on $\Lambda_0$ is easily enforced by optimizing instead over $L_0$, the Cholesky decomposition ($\Lambda_0 = L_0 L_0^T$). By using this Cholesky decomposition and substituting in the equality constraints, (\ref{eq:opt_prob}) is made unconstrained, and may be solved by stochastic gradient descent. Note that in practice, rather than evaluating the loss only on $(x_{t_j+1}, y_{t_j+1})$, we minimize the average loss evaluated on $\{(x_t,y_t)\}_{t=t_{j}+1}^\tau$.

\paragraph{Online Phase.}

\begin{algorithm}[t]
\caption{\label{alg:alpaca_on} \algName{}: Online}
\centering
\begin{algorithmic}[1]
\Require prior terms $\Lambda_0, \Kb_0, \w$, noise variance $\Sigma_\epsilon$, data $\{(x_t,y_t)\}_{t=1}^\tau$
    \State $Q_0 \gets \Lambda_0 \Kb_0$
    \State Compute $\Lambda_0^{-1}$
    \For{$t = 1, \ldots, \tau$}
        \State $\Lambda_t^{-1} \gets \Lambda_{t-1}^{-1} - \frac{1}{1 + \phi_{t}^T \Lambda_{t-1}^{-1} \phi_t} (\Lambda_{t-1}^{-1} \phi_t) (\Lambda_{t-1}^{-1} \phi_t)^T$
        \State $\Q_t \gets \phi_t y_{t}^T + \Q_{t-1}$
        \State $\Kb_t \gets \Lambda_t^{-1} Q_t$
        \State $\bar{y}_{t+1} \gets K_t^T \phi_{t+1}$
        \State $\Sigma_{t+1} \gets (1 + \phi^T_{t+1} \Lambda^{-1}_t \phi_{t+1}) \Sigma_\epsilon$
    \EndFor
    \State \Return $\bar{y}_{t+1}, \Sigma_{t+1}$
  \end{algorithmic}
\end{algorithm}

 After learning the prior parameters $(\Kb_0, \Lambda_0)$ and the weights for the basis functions $w$, \algName{} computes the online posterior by performing Bayesian linear regression using the data observed online $\DB$. 
Since the number of basis functions used may be large for complex functions, we introduce a recursive Bayesian update that improves computational efficiency. This result is well known in linear system identification and online regression \cite{ljung1983theory}. Let $\Q_0 \coloneqq \Lambda_0 \Kb_0$. Then, the online updates may be written as 
\begin{align}
    \Lambda_t^{-1} &= \Lambda^{-1}_{t-1} - \frac{1}{1 + \phi_{t}^T \Lambda_{t-1}^{-1} \phi_t} (\Lambda_{t-1}^{-1} \phi_t) (\Lambda_{t-1}^{-1} \phi_t)^T\\
    \Q_t &= \phi_t y_t^T + \Q_{t-1}
\end{align}
where $\phi_t = \phi(x_t)$. The first update comes from the Woodbury identity \cite{ljung1983theory}. Importantly, this recursive update rule decreases the complexity from the original formulation's $\mathcal{O}(n_{\phi}^3)$ (resulting from inverting $\Lambda_n$ directly) to $\mathcal{O}(n_{\phi}^2)$. The full recursive online algorithm is presented in Algorithm \ref{alg:alpaca_on}.

Note that $\phi$ is not updated online, in contrast to some other approaches to meta-learning such as MAML \cite{finn2017model}. Moving from gradient updates to least squares yields a simpler algorithm online, which is useful for debugging or verifying autonomous systems. 
Because \algName{} performs GP regression in the weight space as opposed to computing the kernel functions, it achieves linear complexity in the amount of data. In particular, for $n$ context data points and $m$ test data points, \algName{} has complexity $\mathcal{O}(n+m)$. One interpretation of \algName{} is as an approximation method for kernel-based GP regression, similar to randomized kernels \cite{rahimi2008random}. However, instead of randomizing basis functions to approximate well-known kernels, we compute basis functions to maximize predictive power with respect to the prior implicit in the training data.


\section{Related Work}

\label{sec:rw}
Thus far we have introduced \algName{} in the context of GP regression. In this section, we discuss connections to other related works in meta-learning and sample efficient regression.

\paragraph{Meta-Learning.} In this work, we present a Bayesian perspective on meta-learning. Recently, model-agnostic meta-learning (MAML) presented a simple approach to meta-learning with impressive performance. MAML was originally introduced from the perspective of learning an initial set of neural network weights that could efficiently reach the desired weights for a task \cite{finn2017model}, given a handful of gradient steps on samples from that task. Subsequent work then recast this as a form of hierarchical Bayes \cite{grant2018recasting}. Specifically, indexing tasks with $j$, the marginal likelihood of the data $X$ is written
\begin{equation*} p(X\mid w) = \prod_j \left( \int p(x_{j_1}, \ldots, x_{j_\tau} \mid w_j') p(w_j' \mid w) dw_j'\right), 
\end{equation*}
where $x_{j_t}$ denote data points from within task $j$, $w_j'$ denotes the post meta-update weights for task and loss function $j$, and $w$ denote the initial weights. 
They note that MAML estimates the likelihood of the data drawn from task $j$ with a point estimate of $w_j'$, computed via gradient descent. In the case of linear regression, Santos \cite{santos1996equivalence} showed that early stopping of gradient descent to compute $w_j'$ results in a Maximum a Posteriori (MAP) estimate of $w_j'$, regularized by $\| w_j' - w\|^2_Q$, where $Q$ depends on the gradient step size and the number of iterations. In contrast to these formulations, we assume a delta prior on the network weights, and so the weights are not updated online. Additionally, our formulation of a Bayesian last layer allows the posterior (given the context data) to be computed directly. Because we perform this analytical Bayesian update, we do not suffer the effects of the regularizing term, and instead may exactly compute the Gaussian approximation of the posterior. Generally, in the case of linear filtering, recursive least squares approaches have been shown to outperform gradient-based approaches \cite{cioffi1984fast}. These results from the linear case may explain in part why \algName{} makes more efficient use of the online data provided, and can perform faster updates.

The interpretation of MAML as hierarchical Bayes allowed the authors to incorporate Laplace approximation into the weight update \cite{grant2018recasting}, and the authors refer to the resulting algorithm as LLAMA. This approximation results in the point estimate of the updated weights being replaced with a second-order approximation of the posterior. Thus, sampling based methods can be used to approximate the posterior distribution of the data. More recently, PLATIPUS \cite{finn2018probabilistic} and Bayesian MAML \cite{kim2018bayesian} have used variational approaches to move from point estimates to Bayesian posteriors. We wish to emphasize that these approaches are capable of capturing a larger family of distributions (such as multimodal distributions), while we achieve improved sample efficiency and better efficiency in terms of computation. We believe these trade-offs are justified and necessary in many robotics applications. 

\paragraph{Efficient Stochastic Process Regression.} In this work we have considered meta-learning for GP regression. A relaxation of this problem formulation to estimation of general stochastic processes was recently presented in \cite{np2018}. This approach is based on amortized variational inference (as in PLATIPUS \cite{finn2018probabilistic}). In particular, it is a form of conditional variational autoencoder (CVAE) \cite{kingma2013auto,yan2016attribute2image}, where the conditioning variable is a function of the context data. As with other autoencoder-based approaches, this method requires sampling-based estimates of the moments of the distribution. In contrast, in \algName{}, mean and variance may be computed analytically. 
Moreover, \cite{np2018} was only tested on problems with two dimensional output spaces, and it is unclear how well it performs for larger problems, whereas \algName{} performs well even for high dimensional systems (such as the 12D hopper system). Within the context of meta-learning for GPs, \cite{yu2005learning} took a similar approach to the one presented here. However, while they are learning hyperparameter priors based on a distribution over tasks, they do not jointly learn a set of basis functions capable of accurately representing posterior distributions. 

\paragraph{Bayesian Output Layer Models.} The approach of affixing a Bayesian last layer to a standard neural network has seen application across machine learning. In \cite{snoek2015scalable}, the authors use this network structure with an arbitrary prior for Bayesian optimization, and \cite{azizzadenesheli2018efficient} use it to improve exploration efficiency in reinforcement learning. In the context of meta-learning, \cite{bauer} use a similar approach to the one presented here. However, they first train their network without considering training it with respect to meta-updated last layers and losses, as we have here. They then perform meta-learning by updating the last layer using approximate methods. In contrast to this approach, we learn network weights such that they form useful basis functions for estimating the posterior, for any amount of data accrued online. Our work also shares common features with recent work on meta-learning with closed form solvers \cite{bertinetto2018meta}, in which the authors do meta-learning with a ridge regression last layer. Generally speaking, these approaches all make arbitrary assumptions regarding the output layer prior, whereas we map our prior information (available via the training data) to a prior over weights. 

\section{Experiments}

\begin{figure}[t]
    \centering
    \includegraphics[width=\columnwidth]{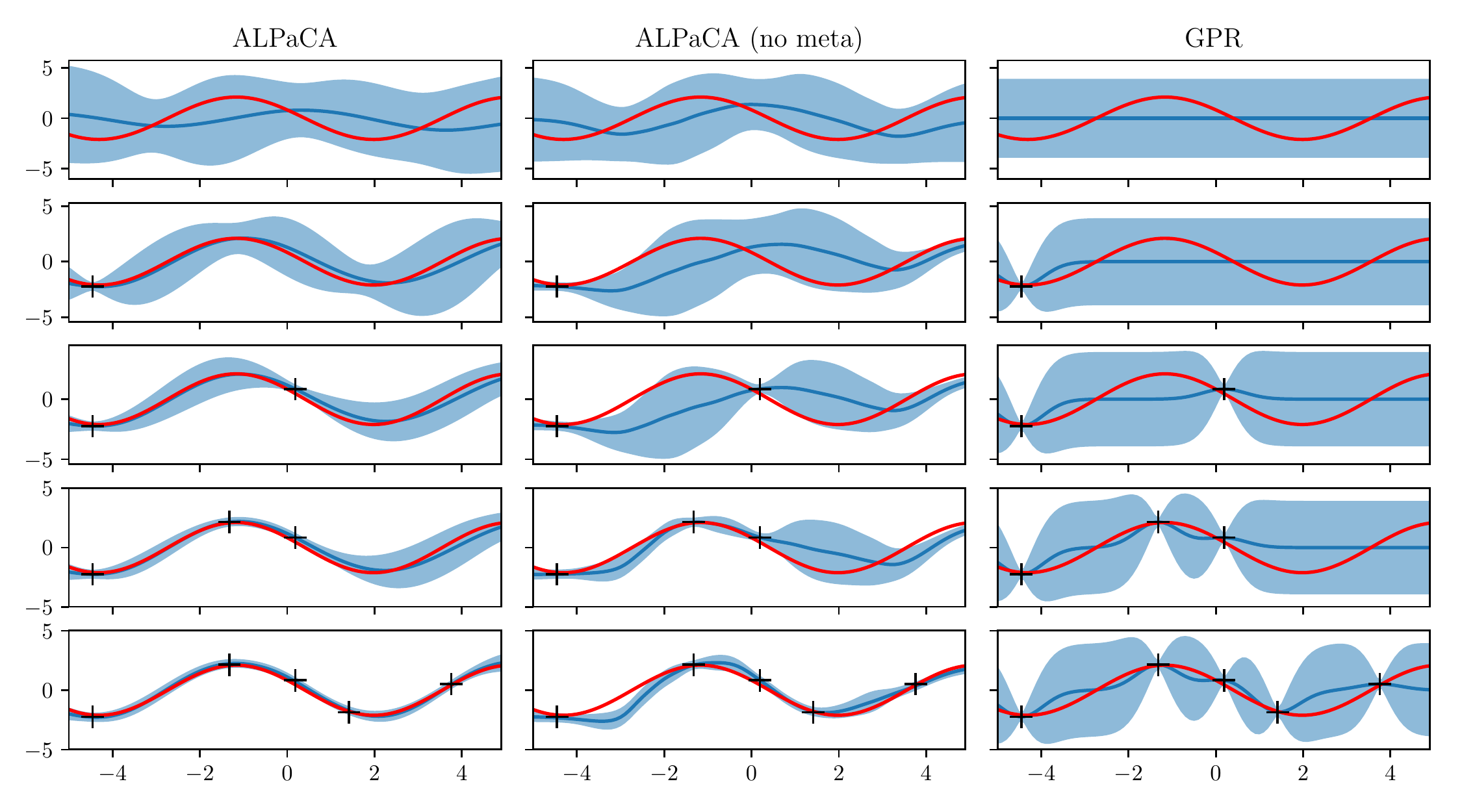}
 \caption{Comparison of \algName{} (left) compared to \algName{} without meta-training (center) and Gaussian process regression (right) on a simple sinusoid problem from \cite{finn2017model}. The plots on each row contain 0, 1, 2, 3, and 5 context samples respectively. Confidence intervals are 95\%.}
    \label{fig:sin}
\end{figure}

\begin{figure}[t]
    \centering
    \includegraphics[width=\columnwidth]{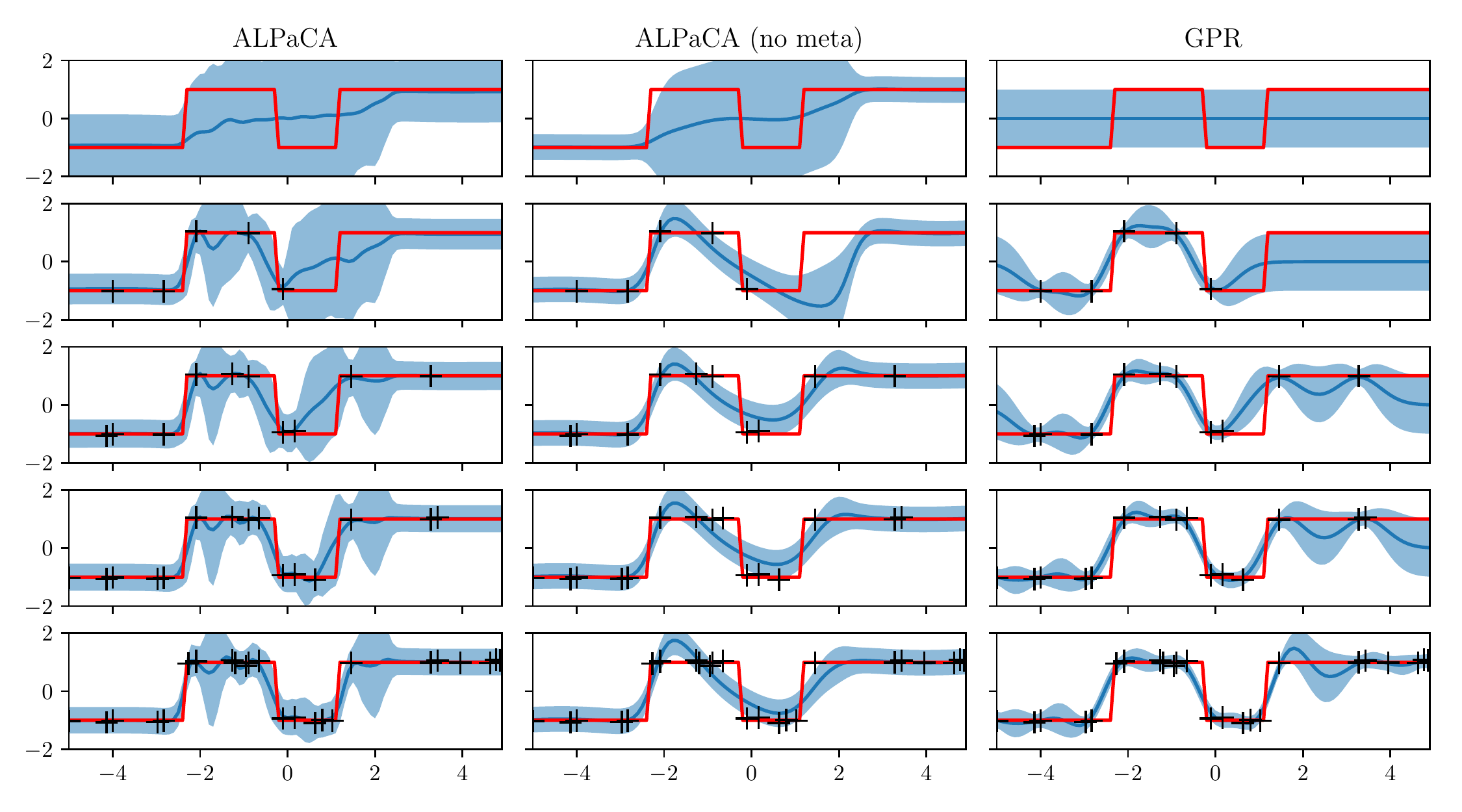}

    \caption{Comparison of \algName{} (left) compared to \algName{} without meta-training (center) and Gaussian process regression (right) on a function exhibiting discrete switching between $y=-1$ and $1$ for $x\in[-2.5,2.5]$. Three switching points are randomly sampled. The plots on each row contain 0, 5, 10, 15, and 25 context samples.}
    \label{fig:step}

\end{figure}

In this section, we investigate the performance of \algName{} on a variety of regression problems, ranging from simple toy problems which help in building intuition for the algorithm, to challenging prediction and dynamics modeling problems. 
We compare \algName{} against several other algorithms that are capable of rapid online adaptation. First, as an ablation experiment, we compare to \algName{} without the meta-training. This is to say, the network is trained to predict the prior for varying datasets, but is not explicitly conditioned on random selections of data. This is roughly equivalent to standard methods for training neural networks with Bayesian last layers \cite{snoek2015scalable,azizzadenesheli2018efficient}. In these works, a network is typically trained to minimize e.g. MSE loss over all of the data, and then the prior over the last layer is chosen arbitrarily.
For the dynamical systems that we investigate, we also compare to the predictive performance of \algName{} with no online updating. This network is just trained in expectation over the sampled values of the model parameters $\theta$.
For our toy examples, we compare to GP regression. We chose a zero-mean GP with a squared-exponential kernel, which is often the default choice in GP modeling \cite{snoek2012practical}. The three above approaches are compared in terms of negative log likelihood to compare the accuracy of their posterior predictions. 
In our experiments we use \textit{tanh} nonlinearities, due to favorable properties in terms of variance smoothness and the behavior of the variance far from the observed data. This is discussed in more detail in the appendix.

In addition to these comparisons, we also compare to MAML \cite{finn2017model}, which has rapidly become one of the most common approaches for meta-learning. Indeed, MAML was recently used for online system identification \cite{clavera2018learning} on relatively complex dynamical systems. Because MAML generates point predictions rather than full posteriors, we compare in terms of MSE loss (for \algName{}, we compute MSE loss on the mean of the posterior). We were not able to compare to PLATIPUS \cite{finn2018probabilistic} or Neural Processes \cite{np2018} because to our knowledge, at the time of writing, the code is not publicly available. Throughout our experiments, plotted confidence intervals are 95\%.



\subsection{Toy Experiments}

We first investigate two simple regression examples that allow both quantitative and qualitative evaluation of performance. First, we investigate a regression problem over a family of sinusoids, where the amplitude and phase were sampled uniformly from $[0.1,5.0]$ and $[0,\pi]$ respectively. This sinusoid problem was used as a simple test problem in \cite{finn2017model}.
We also evaluate performance on a new, simple benchmark that consists of a family of two-valued functions that begin at $y=-1$ for $x<2.5$, end at $y=1$ for $x>2.5$, and switch three times between these values for $x\in[-2.5,2.5]$.
These switching points are sampled uniformly from this range. This function was designed to be challenging relative to its apparent simplicity for our approach. Because we are performing basis function regression, the discrete switches between $-1$ and $1$ which can occur anywhere within the range specified above, are hard to capture. This is also challenging for GP approaches, as the structure of the function is hard to capture via a single global smoothness parameter in the kernel.
For example, in \cite{SvenssonSchon2017}, the authors learn separate GPs for each piecewise continuous segment, as opposed to loosening their smoothness priors.

The outputs of \algName{}, \algName{} without meta-training, and Gaussian process regression (GPR) on these two problems are plotted in figures \ref{fig:sin} and \ref{fig:step}. 
First, note that \algName{} is capable of making predictions that incorporate structural knowledge of the problem from a single datapoint, compared to the local predictions that are common for GPs with the SE kernel. Secondly, on the sinusoid problem, the first sample allows \algName{} to correctly reduce uncertainty at points at a distance of a scalar multiple of a half wavelength from the sampled point. Moreover, one sample is enough to have a good estimate of the sinusoid nearly everywhere, and within five samples the estimated variance has dropped to nearly zero. This reduction in variance is observed for \algName{} without meta-training as well, but its predictions are incorrect, highlighting the utility of our meta-learning approach. Further experiments with varying noise covariance are presented in the appendix.

\begin{figure}[t]
    \centering
    \includegraphics[trim={0 0 0 1cm},clip,width=0.20\columnwidth]{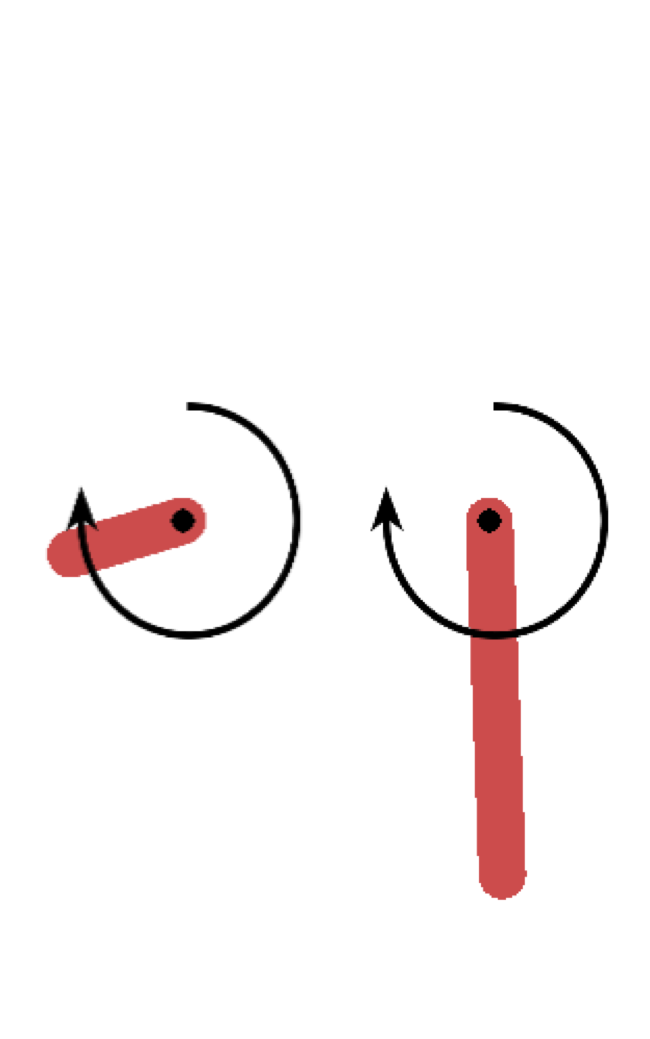}
    \includegraphics[width=0.39\columnwidth]{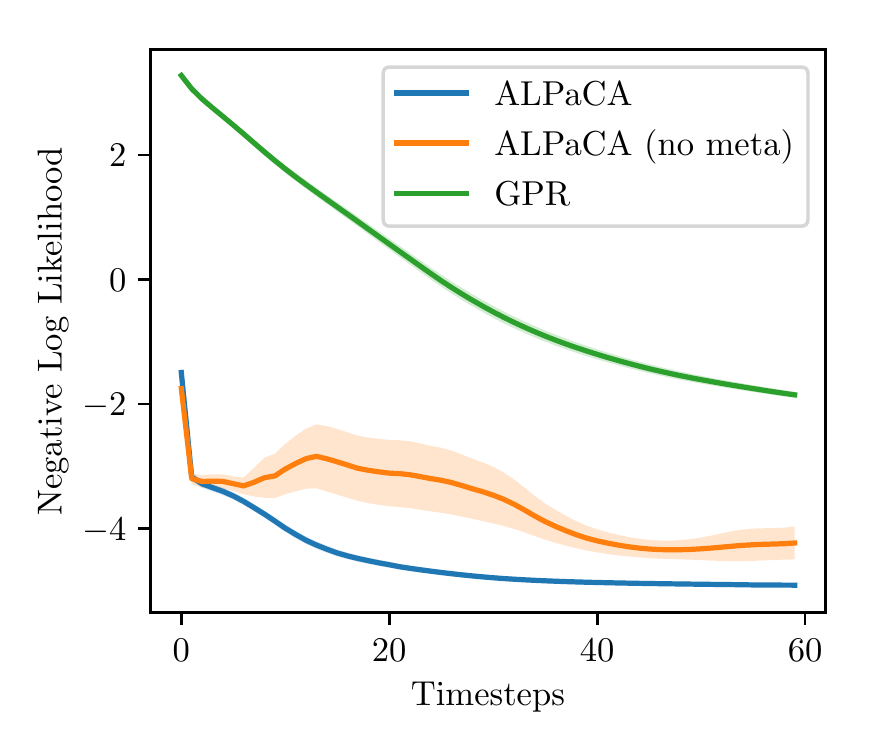}
    \includegraphics[width=0.39\columnwidth]{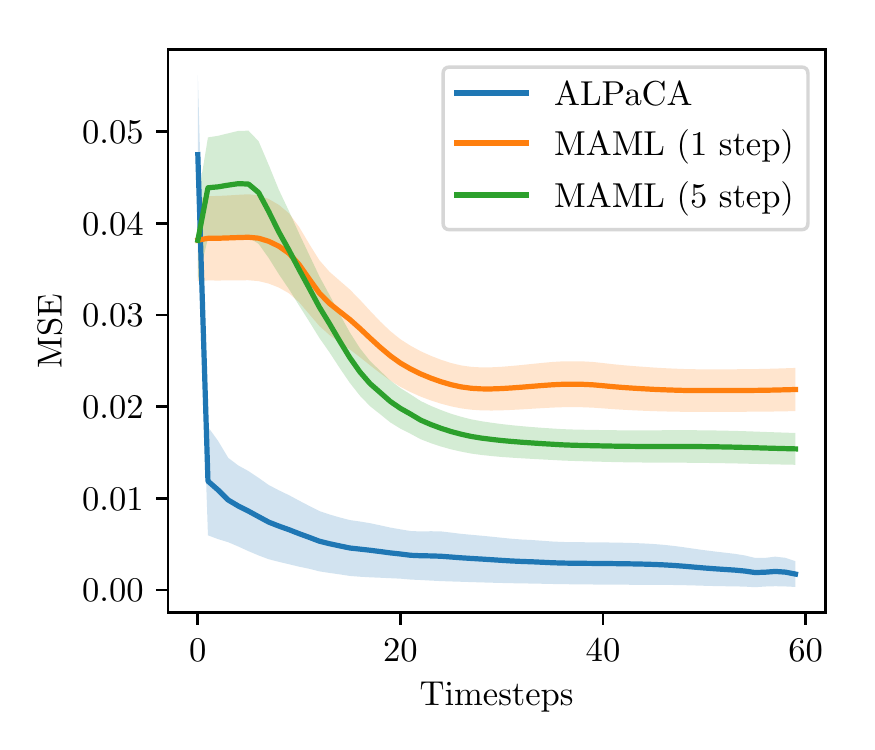}

    \caption{Negative Log Likelihood (center) and Mean Squared Error (right) for the pendulum system. The image on the left shows the limits of the uncertainty over the pendulum length.}
    \label{fig:nll_pend}
\end{figure}

On the step function, \algName{} can capture the discrete steps as well as or better than GP methods. Indeed, \algName{} does not exhibit the large overcorrection below the function that GPR exhibits. Again, \algName{} without meta-training reduces the predicted variance incorrectly. In contrast, the predicted confidence intervals generated by \algName{} appear to be well-calibrated. \algName{} with and without meta-training successfully learn that the function always takes a value of $-1$ and $1$ for $x$ below $-2.5$ and above $2.5$ respectively. This automatic incorporation of prior information is a powerful tool in the context of robotics, where a wealth of physical information is available but often hard to encode into standard GP regression.

The average negative log likelihood  and the mean squared error (based on the mean prediction) of the test set as a function of the number of context samples is plotted for both problems in the appendix. We find that while the prior (no samples observed) for \algName{} and \algName{} without meta-training is nearly identical, \algName{} nearly monotonically improves in performance as more samples are gathered online. We wish to note that GPR performance could likely be improved by better choice of prior, but as we discussed above, this tuning of the prior is often difficult for complex systems. We also compare \algName{} to MAML in terms of MSE. We find that we outperform MAML for all context data sizes, but this difference in performance is especially acute for a small number of context datapoints. Indeed, because MAML performs gradient steps on context data, it performs poorly with a single sample.

\subsection{Dynamical Systems}

We investigate the performance of \algName{} in fitting the transition model for dynamical systems. In these tasks, we wish to fit a model to use the input $x=(s_t,a_t)$ to predict $y=s_{t+1} - s_t$, where $(s_t, a_t, s_{t+1})$ represents the state, action, and next state of a dynamical system. The process generating these $(x,y)$ pairs is given by the true dynamics of the system, $s_{t+1} - s_t = f(s_t,a_t;\theta)$, where $\theta$ here represents model parameters. We generate a dataset of trajectories under different settings of $\theta$ sampled from a prior, and study whether \algName{} can capture this structured uncertainty to enable sample efficient Bayesian regression online.

\begin{figure}[t]
    \begin{subfigure}{0.2\textwidth}
    \centering
    \includegraphics[width=\columnwidth]{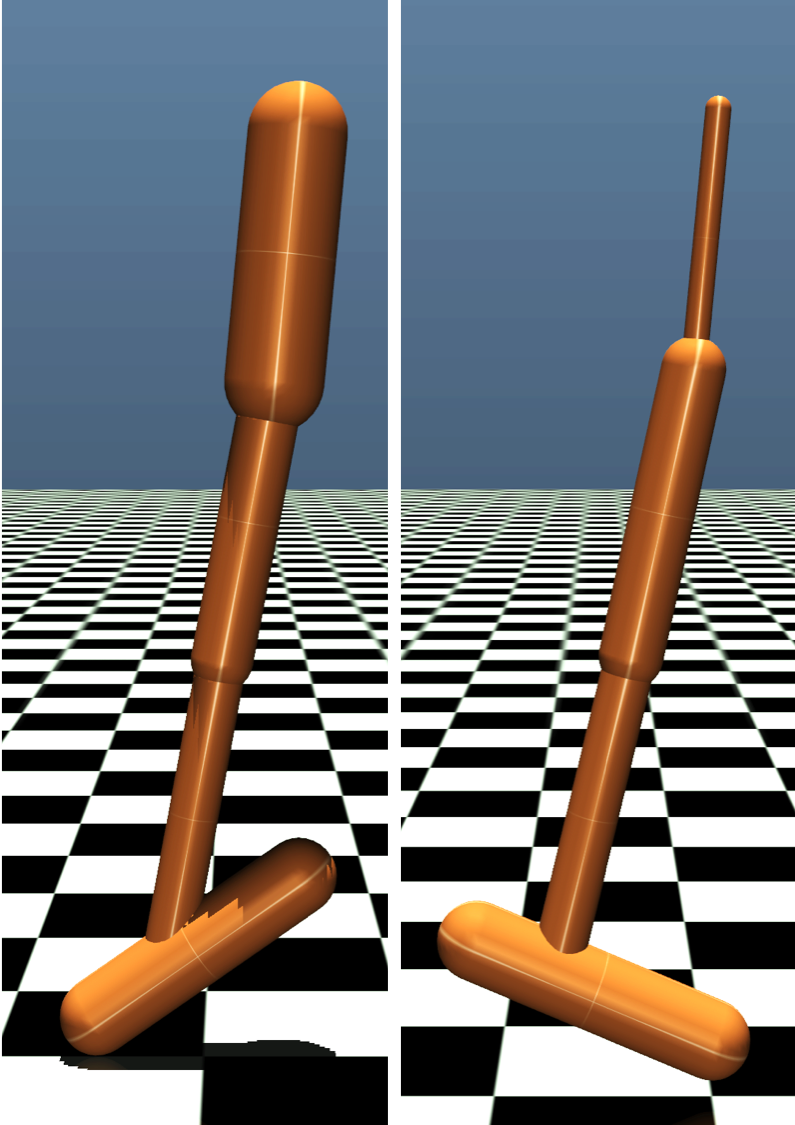} 
    \end{subfigure}
    \begin{subfigure}{0.39\textwidth}
    \includegraphics[width=\columnwidth]{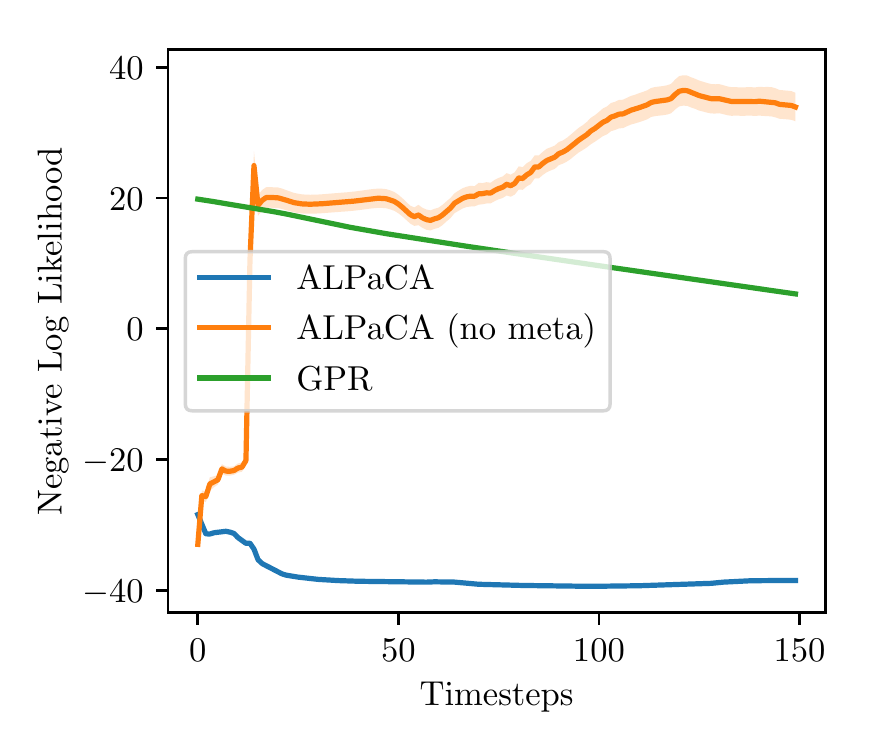}
    \end{subfigure}
    \begin{subfigure}{0.39\textwidth}
    \includegraphics[width=\columnwidth]{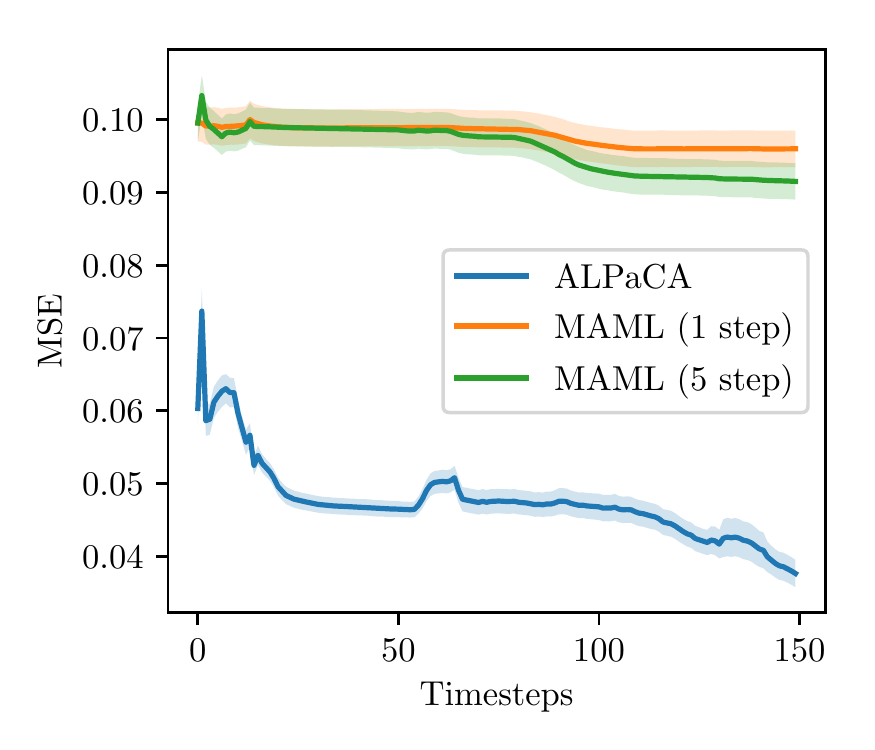}
    \end{subfigure}
    
\caption{Negative Log Likelihood (center) and Mean Squared Error (right) for the hopper system. The image on the left depicts bounds on the uncertain torso size of the hopper.}
    \label{fig:nll_hop}
\end{figure}

We test \algName{} on the pendulum and the hopper from OpenAI's Gym \cite{brockman2016openai}. For these examples, the dynamics parameters were varied. For the pendulum, the mass and length were sampled uniformly from $[0.5,1.5]$. The hopper's foot friction was sampled uniformly from $[1.7,2.0]$ and the torso size was sampled uniformly from $[0.02,0.08]$. Plotted rollouts for these systems are provided in the appendix. 

The performance of \algName{} on the pendulum and the hopper are plotted in figures \ref{fig:nll_pend} and \ref{fig:nll_hop}. Again, when the amount of context data is low, MAML provides little performance increase. \algName{}, in contrast, achieves rapid performance improvement within the first few samples. These results also highlight an issue with approaches such as MAML, where taking more gradient steps can lead to overfitting when the amount of context data is small. The selection of number of gradient steps to take with a MAML-style approach, as well as the batch size to be used for online updates, are hyperparameters that are relatively difficult to tune. By computing the posterior directly via recursive least squares, \algName{} avoids this ambiguity in the number of gradient steps to take.
Kernel-based methods for GP regression are known to be prohibitively slow for high dimensional dynamical systems; the time required to evaluate \algName{} is compared to GPR are plotted and discussed in the appendix.



\subsection{Vehicle Lane Change}

\begin{figure}[t]
    \centering
    \includegraphics[width=0.20\columnwidth]{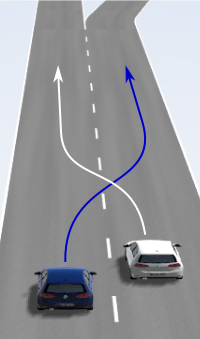}
    \includegraphics[width=0.39\columnwidth]{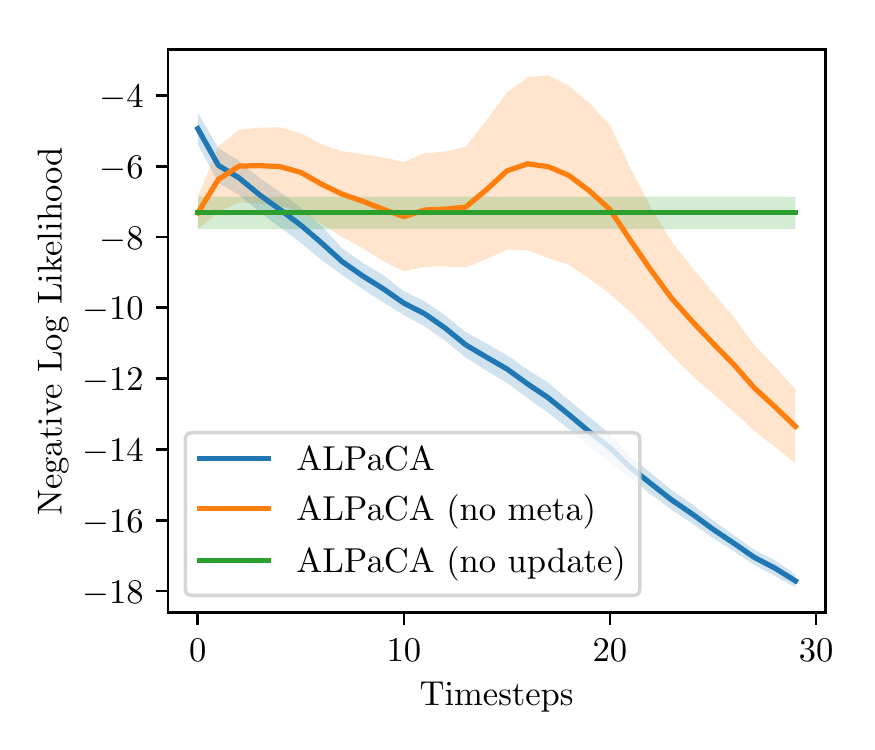}
    \includegraphics[width=0.39\columnwidth]{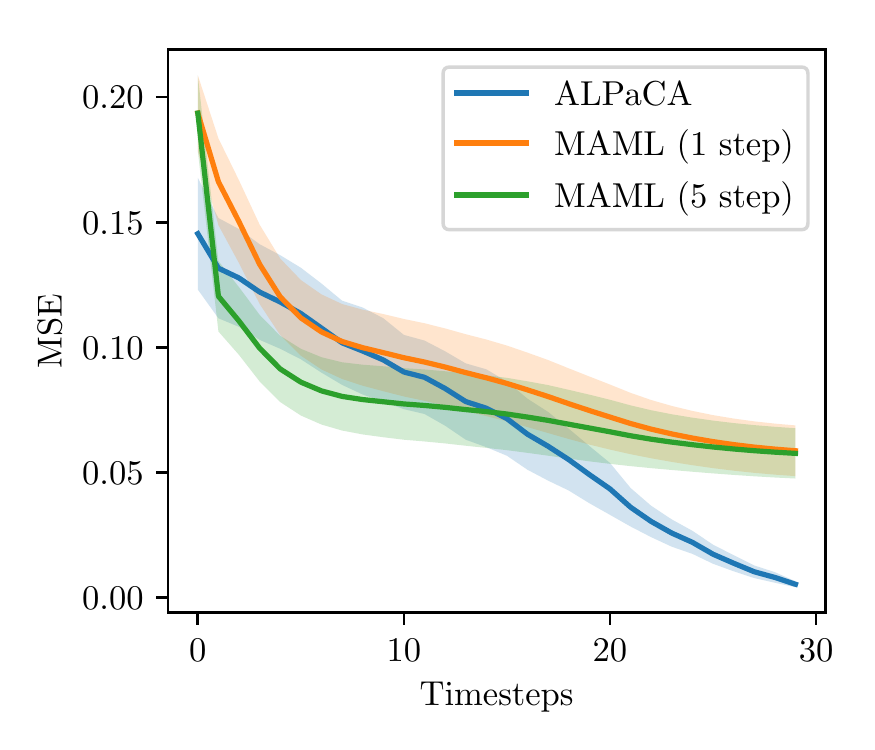}
    \caption{Negative Log Likelihood (center) and Mean Squared Error (right) for the lane change experiment. The image on the left shows the view for the drivers, and the aim of the task (reproduced with permission from \cite{schmerling2017multimodal}).}
    \label{fig:nll_lane}
\end{figure}

Finally, we investigate the performance of \algName{} on human lane change experiments\footnote{Data available at: https://github.com/StanfordASL/TrafficWeavingCVAE}. These experiments were based on the dataset presented in \cite{schmerling2017multimodal}. This data is based on 19 pairs of participants, driving against each other. The objective of the task was to switch lanes with each other without communicating verbally, and without colliding. Further details on use of this dataset are presented in the appendix. Here, we treat the system as an autonomous dynamical system, and use \algName{} to predict the transitions. 

In this experiment, the underlying parameter $\theta$ is essentially capturing the driving style of the human. \algName{} aims to identify the underlying style of the drivers from observations of the first few transitions of a particular trial to reduce uncertainty over future predictions. Numerical results are presented in figure \ref{fig:nll_lane}. Interestingly, the rate of performance improvement on this problem was slower than previous problems. On previous problems, rapid performance improvement occurred early in the episode, which tailed off toward the end of the episode. A possible explanation is that the transition data is often bimodal, with a mode corresponding to driving straight and a mode corresponding to switching lanes. In these experiments, \algName{} performance improves nearly linearly. Surprisingly, MAML does not experience an initial performance decrease as in previous experiments, and shows rapid improvement that quickly slows. 

While building models to identify dynamics parameters as above is reasonably simple, identifying latent variables such as driver intent and style are difficult to do via non-data-driven models. 
Critically, we did not specify in any way that we wish to identify human driving behavior over the course of the episode. Generally, pre-specification of the latent variables to identify is not necessary. \algName{}, if trained from real operational data, will learn to identify existing latent parameters. 


\section{Discussion and Conclusions}

In this work we have presented \algName{}, a sample efficient probabilistic neural network model that enables efficient adaptive online learning. We have demonstrated that \algName{} outperforms kernel GP regression, which is a tool used throughout robotics and engineering as a whole, as well as MAML \cite{finn2017model} (which itself has outperformed many meta-learning approaches). In contrast to kernel GP regression, our approach enables incorporation of an arbitrarily large amount of prior information, and scales to large data regimes. In contrast to MAML, \algName{} maintains a full analytic posterior, which is useful for applications such as Bayesian optimization and stochastic optimal control. Moreover, it is simple to implement and a plug-and-play tool for any regression problem. 
We believe its blend of high-capacity regression models, together with sample efficiency, has the potential to improve robotic autonomy in many scenarios.

\paragraph{Limitations.} While the \algName{} model has shown potential for efficient online Bayesian learning, it has several limitations. First and foremost, the use of analytic update formulas come with the standard limitations of Gaussian models -- in particular, the assumption of Gaussian noise. Indeed, in this paper we assume $\Sigma_\epsilon$ is know. We discuss the case where this assumption is relaxed in the appendix. Relative to MAML-style models \cite{finn2017model}, in which all parameters of the model are being adapted online, \algName{} has lower model capacity. Our experiments demonstrate that the reduction in capacity results in an improvement in stability (via near-monotonic MSE improvement relative to MAML), but it may be possible to achieve both of these objectives. Finally, fundamental machine learning considerations such as data requirements and meta-overfitting have largely not been addressed in this paper, or in the meta-learning literature broadly. 

\paragraph{Future Work.} The literature on kernel methods for GP regression is large, and it was impossible to compare to all of the flavors of GP methods that are used in practice. We believe further comparisons of the relative merits of \algName{} and GP methods are useful future work. Beyond this, several avenues of future work exist. We detail them here briefly, and an extended discussion is presented in the appendix. First, because GPs are closed under addition, our method may be combined with kernel GP methods. Additionally, \algName{} is simply extended to tracking time-varying $\theta$ using forgetting methods, which we have not discussed in this work. Generally speaking, many of the tools previously developed within basis function system identification may be applied with \algName{}. Finally, characterization of the data requirements of meta-learning algorithms, as well as meta-overfitting are needed.


\section*{Acknowledgments}

This work was supported by the Office of Naval Research YIP program (Grant N00014-17-1-2433), by DARPA under the Assured Autonomy program, and by the Toyota Research Institute (“TRI”).
This article solely reflects the opinions and conclusions of its authors and
not ONR, DARPA, TRI or any other Toyota entity. James Harrison was supported in part by the Stanford Graduate Fellowship
and the National Sciences and Engineering Research Council (NSERC).



\bibliography{cites}
\bibliographystyle{ieeetr}

\clearpage
\appendix


\section{Further Algorithmic Details and Extensions}

\subsection{Choice of Nonlinearity}

The choice of neural network nonlinearity strongly affects the behavior of the variance far from datapoints. Moreover, choice of smooth vs. non-smooth activation affects the smoothness of the mean, but has an especially prevalent effect on the smoothness of the variance. In this work we have chosen to use the \textit{tanh} activation function, which is bounded. As a result, and contrary to what is observed with a \textit{ReLU} activation, the variance far from data reaches some prior value, similar to the behavior of a GP. This same design choice was discussed by \cite{snoek2015scalable} and \cite{gal2016theoretically}. Whether or not very large variance far from data is desirable is left to the system designer. 

\subsection{Dependent Samples}

We have previously assumed independence of data $x$ when conditioned on $\theta$. In the case of dynamical systems, this is not true. Generally, the data generating process may be written as a hidden Markov model (HMM), in which the underlying state, $x$, has Markovian dynamics, and we receive observations $y$. In this case, we use the term HMM to refer to a process that has a possibly continuous state space, as opposed to more common usage which assumes a finite state space. 
In this case, the joint distribution of the dataset and the current $x, y$ data and label pair may be written
\begin{equation*}
    p(x_t,y_t,\mathcal{D} \mid \theta) = p(x_0 \mid \theta) \prod_{i=1}^t p(y_i \mid x_i, \theta) p(x_i \mid x_{i-1}, \theta).
\end{equation*}
However, in the online case, we do not observe the state and thus computation of the posterior requires that the state be jointly estimated. This problem of simultaneous system identification and state estimation is common in the system identification literature \cite{schon2011system}. In this work, we will assume that the state is fully observed without noise, and thus $y_t = x_t$. The assumption of full state observations is common to make the problem tractable without expensive filtering jointly over the non-Markovian observations and the model parameters \cite{schon2011system}. Alternatively, it is possible with observation noise to simply perform prediction in the space of observations. 

A common method for time series prediction (in which the underlying model is an HMM, but predictions are made for the observation) is an autoregressive approach, in which the next observation is predicted as a function of some finite number of previous observations \cite{hamilton1994time}. Such an approach, which effectively corresponds to concatenating several previous observations before input to $\phi$, may easily be applied in the context of \algName{}. Moreover, a wide variety of approaches within the literature on autoregressive models are likely applicable within the \algName{} framework. Extension to recurrent neural network models (e.g. \cite{hochreiter1997long}) is possible, but this would substantially increase training complexity. Indeed, recurrent models for meta-learning are an existing topic of interest within the meta-learning community \cite{santoro2016meta}, and so combination with those works is possible.




\subsection{Time-Varying $\theta$}

The method we have presented in this work meta-trains basis functions for online Bayesian regression. However, \algName{} is agnostic to the choice of online regression algorithm. As an example of the potential utility of this, consider the case in which $\theta$ is time-varying. In this case, off-the-shelf forgetting algorithms such as \cite{kulhavy1987restricted} may be used for the online phase with no other modifications. There is a vast literature on system identification methods based on regression over basis functions, much of which can be applied within the \algName{} framework. This problem was addressed in the context of meta learning by \cite{al2017continuous}, who used a MAML-based approach that jointly learned an adaptation rate. In contrast, an exponential forgetting method is considerably simpler. 

\subsection{Combination with Existing GP Regression Methods}

One of the primary uses of GP regression in robotics is for algorithms which aim to guarantee safety online \cite{berkenkamp2017safe}. As such, the strong locality of squared exponential kernels is a selling point, and the ability of \algName{} to induce further structure in the regression problem is not desirable. However, it is possible to incorporate SE kernel behavior into \algName{}. Because Gaussian processes are closed under addition, one could combine \algName{} directly with a kernel-based GP model via convex combination of means and variances. This is a slightly unusual model, but provides a spectrum of intermediate models between kernel-based GP regression and \algName{} models. Moreover, if safety is critical, Gaussian models have a variety of model selection algorithms \cite{birge2001gaussian}. These approaches may be used to detect if the predictions from an \algName{} model are far out of distribution (corresponding to a misspecified prior), which can be used to inform weights on GP models. 

\subsection{Known Noise Covariance} 

In this paper we have addressed the problem of Bayesian linear regression with known noise covariance, $\Sigma_\epsilon$. This assumption is in line with the majority of the literature on GP regression. Relaxing this assumption changes the posterior distribution over $\K$, which is a Normal-Inverse-Wishart distribution as opposed to a Matrix Normal. As a result, the posterior predictive density for $y$ is a multivariate $t$ distribution. In the case of a scalar output, this regression problem was discussed in \cite{shah2014student}, however the relative performance improvement of this approach was primarily for noisy systems with a relatively small amount of data. This process has been discussed by numerous others in the statistics and machine learning literature. 

While Bayesian regression with the multivariate $t$ distribution is tractable in relatively simple cases, \algName{} requires backpropagation through the posterior predictive distribution. This distribution involves a gamma function for the univariate case, which may add numerical instability. An approach is that of \cite{parsa}, in which they approximate the gamma with Stirling's approximation. However, the general multivariate $t$ distribution involves the multivariate gamma distribution, for which approximation methods are more difficult. For example, in \cite{SvenssonSchon2017} the authors turn to MCMC methods for identification of the hyperpriors for the Normal Inverse Wishart (equivalent to multivariate $t$ in this case). 

\subsection{Data Acquisition.} There are two approaches to gathering the data $\mathcal{D}$ to train \algName{}. First, given a process wherein $\theta$ is being sampled from $p(\theta)$, and $x,y$ pairs are observed, this data may simply be used for meta training. A simple example is predicting driver actions, wherein one can gather data from observing driver state transitions for different drivers.  Alternatively, with access to $f$ and a belief over $\theta$, one may train \algName{} via sampling $\theta$ from the belief, followed by sequentially sampling $x$ and $y$. This case is more common when the system is relatively well known. For example, this is often the case for robotic system identification, in which the governing dynamics of the system are known but there is uncertainty with respect to the parameters.

In the online/streaming setting, data gathered online may be used to improve the model in multiple ways. First, it may simply be added to the training set, and the model may be retrained. Second, if the training is done via simulation of $f$, online data may be used to update the prior belief over $\theta$ via filtering algorithms that may be too expensive to run online. 

\section{Further Experimental Details and Results}

\subsection{Toy Experiments}

\begin{figure}
    \centering
    \includegraphics[width=\columnwidth]{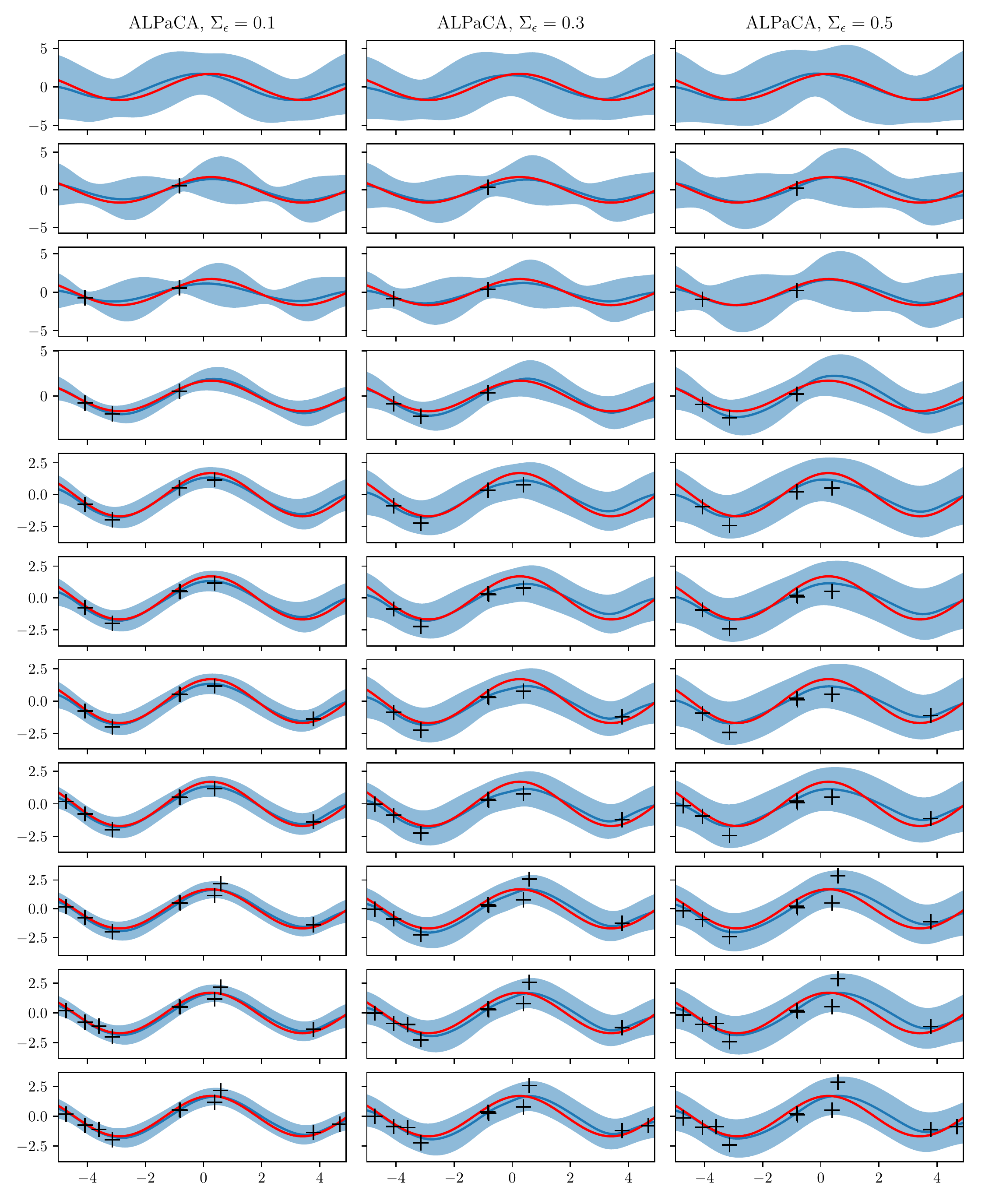}
    \caption{Results for \algName{} for the sinusoid problem with varying $\Sigma_\epsilon$. The noise covariance is fixed for each column, the rows show 0 -- 10 samples.}
    \label{fig:sin_noise}
\end{figure}

\begin{figure}
    \centering
    \includegraphics[width=\columnwidth]{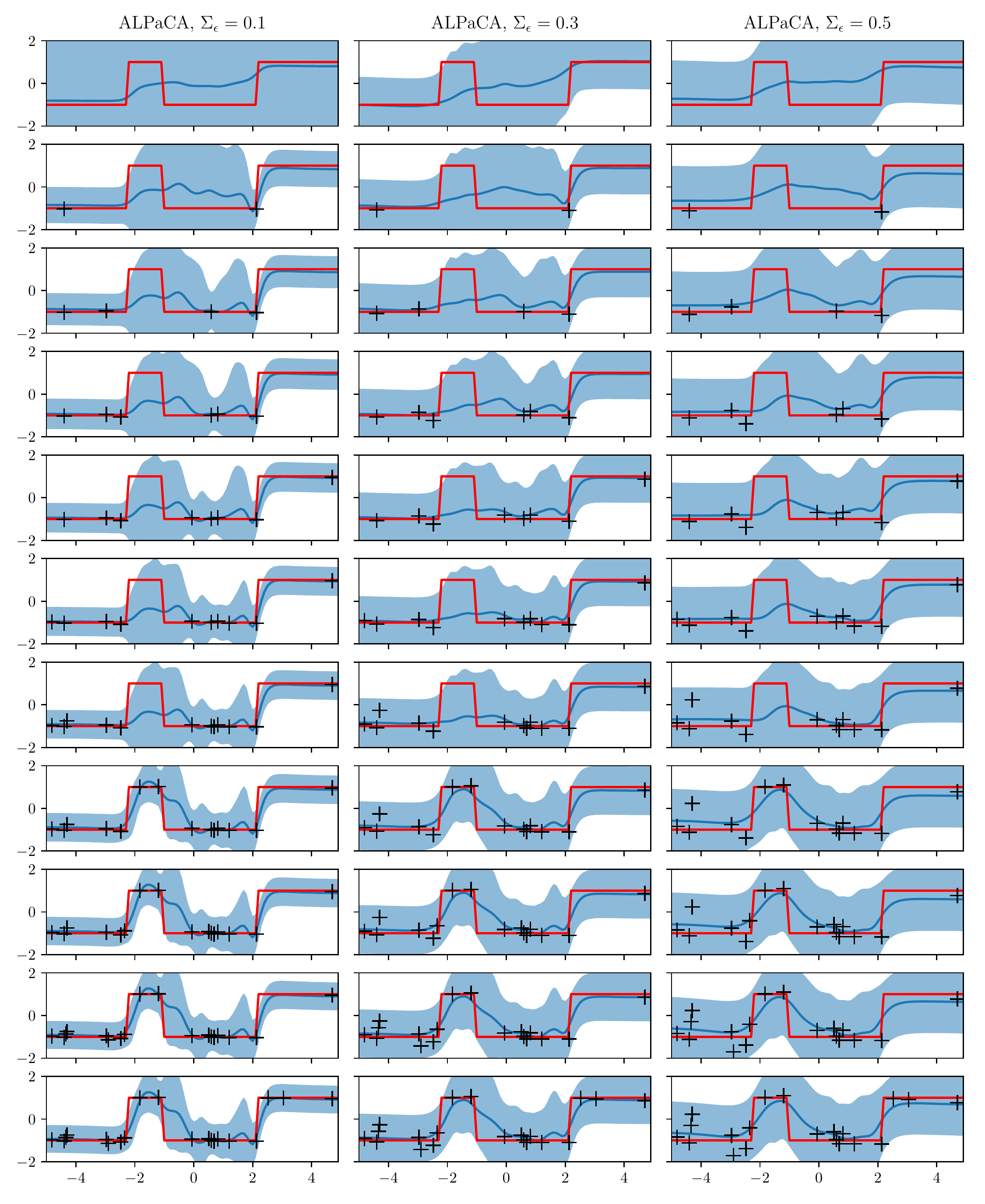}
    \caption{Results for \algName{} for the step function problem with varying $\Sigma_\epsilon$. The noise covariance is fixed for each column, the rows show 0 -- 20 samples, in steps of $2$ samples per row.}
    \label{fig:step_noise}
\end{figure}

For both problems, datapoints are sampled uniformly between $x = -5$ and $5$.  For the sinusoid problem, the network was composed of two hidden layers of $128$ units each, and $16$ basis functions were used. 
For the step function, the same network architecture was used, but we used $128$ basis functions. We did not notice a significant increase in computation time in the online phase for this number of basis functions, but training was slower than for the sinusoid problem. For both problems, $\Sigma_\epsilon = 0.05$. We also investigated both the sinusoid and the step with larger noise values, and these are plotted in figures \ref{fig:sin_noise} and \ref{fig:step_noise}.

In our comparison to MAML \cite{finn2017model}, the performance of MAML (in terms of MSE) was slightly worse than the reported values. However, \algName{} performs better than the values reported in the original paper as well. This mismatch is likely due to differences in network training time and hyperparameter tuning. In this paper we did not perform experiments to optimize hyperparameters. 

\subsection{Dynamical Systems}

The pendulum and hopper systems were based on default implementations in OpenAI's gym \cite{brockman2016openai}. For the pendulum, we use $\Sigma_\epsilon = 0.001 I$, while for the hopper, we use $$\Sigma_\epsilon = \mathrm{diag}([ 10^{-3}, 10^{-4}, 10^{-4}, 10^{-5}, 10^{-7}, 10^{-6}, 10^{-3}, 10^{-3}, 10^{-3}, 10^{-3}, 10^{-3}, 10^{-3}]).$$ This was chosen somewhat arbitrarily, as these problems have deterministic transitions (based on a physics simulator). Intuitively, $\Sigma_\epsilon$ is a hyperparameter that can either be identified from data, or chosen arbitrarily. When chosen arbitrarily, it controls the tradeoff between the variance term and normalized MSE term. For the hopper the different state elements varied with different orders of magnitude, and thus $\Sigma_\epsilon$ was chosen to match. 

\begin{figure}
    \centering
    \includegraphics[width=\columnwidth]{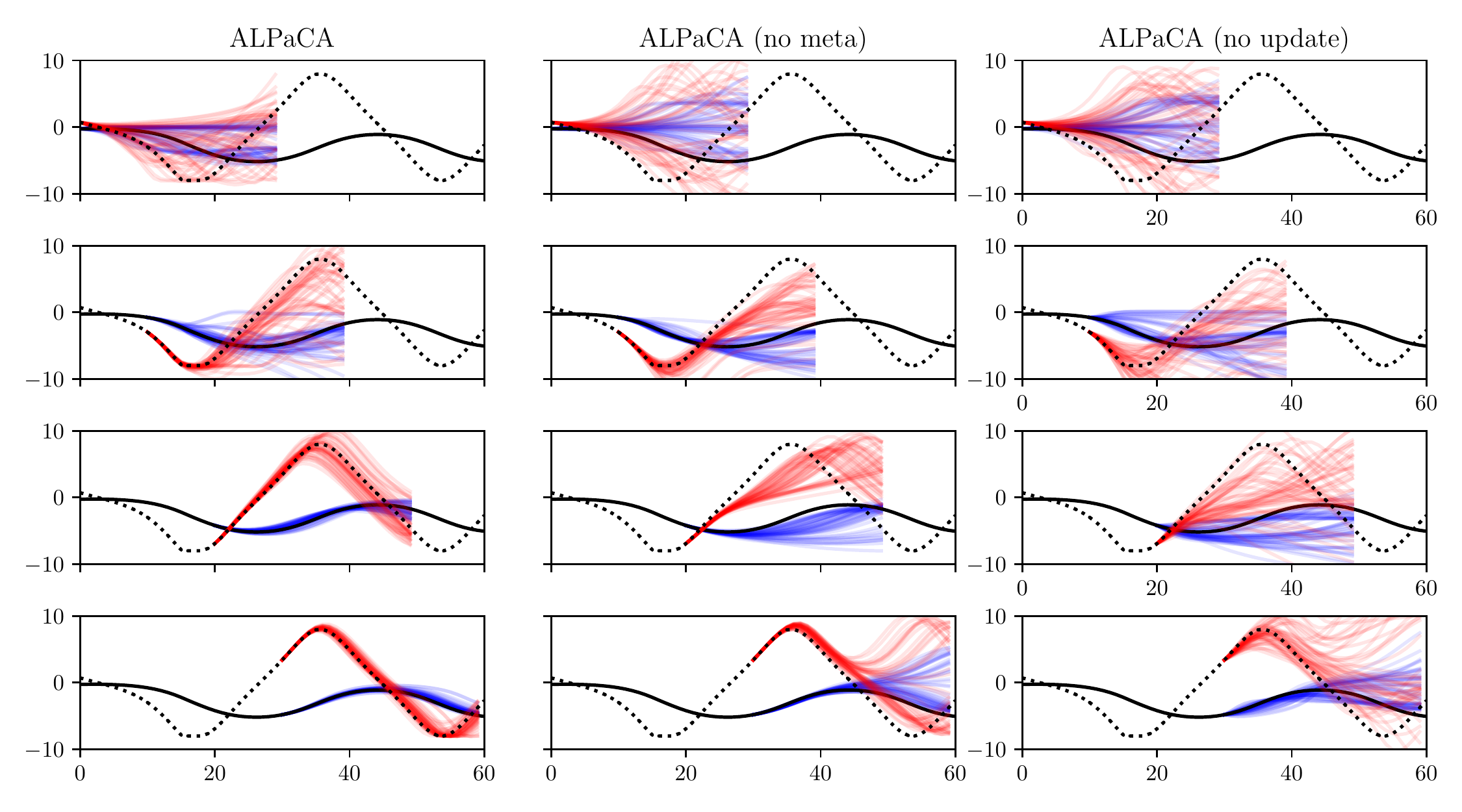}
    \caption{Rollouts for the pendulum problem for \algName{}, \algName{} without meta-updates during training, and \algName{} without meta-updates during execution. Rollouts were generated by sampling $K$ and recursively evaluating the dynamics with the same value of $K$. The two curves (red and blue) are the two elements of the state, and the dotted black and solid black line denote the true rollout.}
    \label{fig:pend_rollout}
\end{figure}

For the pendulum, the system state consists of angle and angular velocity. The angle was sampled uniformly between $0$ and $2 \pi$ at the start of each episode, and the angular velocity was sampled uniformly between $-8$ and $8$ (default initialization). The system was simulated in open-loop (zero action) to generate data. For this problem, two hidden layers of $128$ units each were used, and $16$ basis functions. Rollouts of the system are plotted in figure \ref{fig:pend_rollout}. In this figure, \algName{} is compared to \algName{} with no meta updating during training, and \algName{} with no online updating. Note that \algName{} without online adaptation is roughly equivalent to ``dynamics randomization'' \cite{peng2017sim}, which has become a common tool to increase robustness of learned policies.

\begin{figure}
    \centering
    \includegraphics[width=0.49\columnwidth]{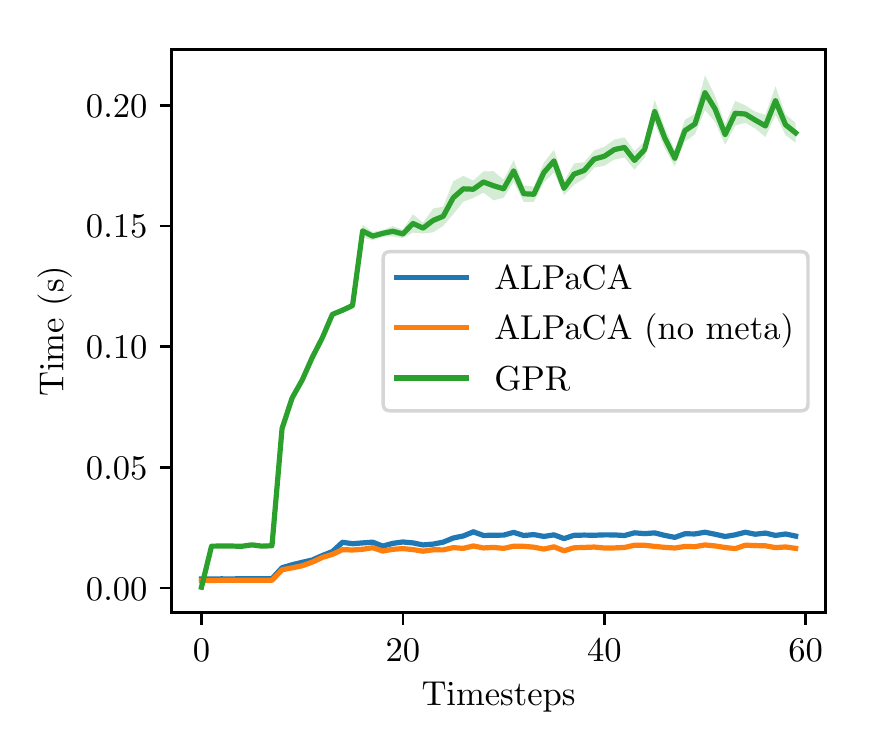}
    \includegraphics[width=0.49\columnwidth]{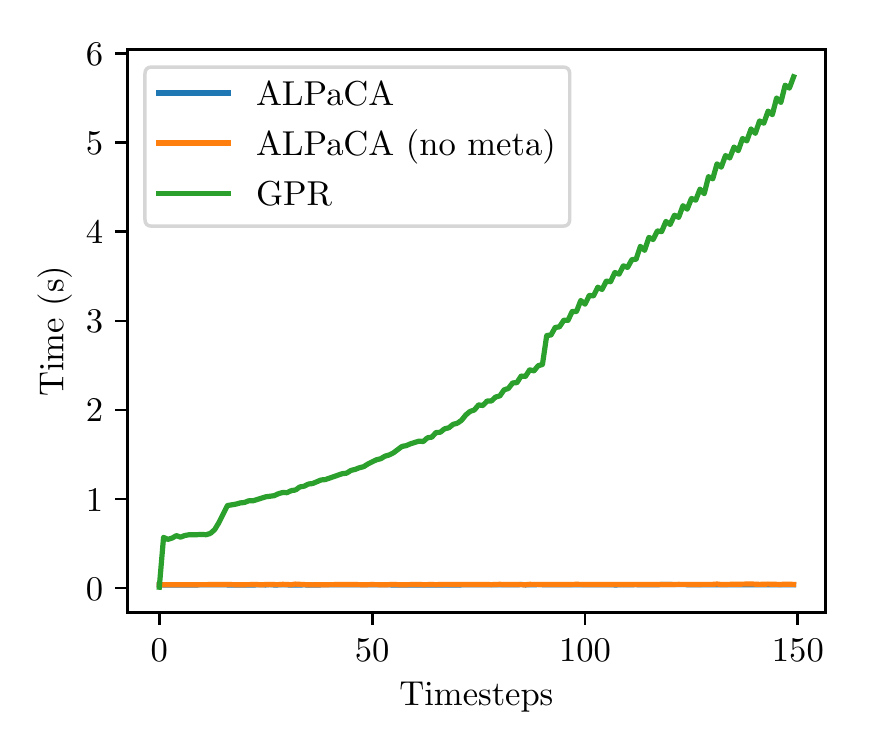}
    \caption{Computation time for GPR and \algName{} on the pendulum environment (left) and the hopper environment (right).}
    \label{fig:time}
\end{figure}

\begin{figure}
    \centering
    \includegraphics[width=\columnwidth]{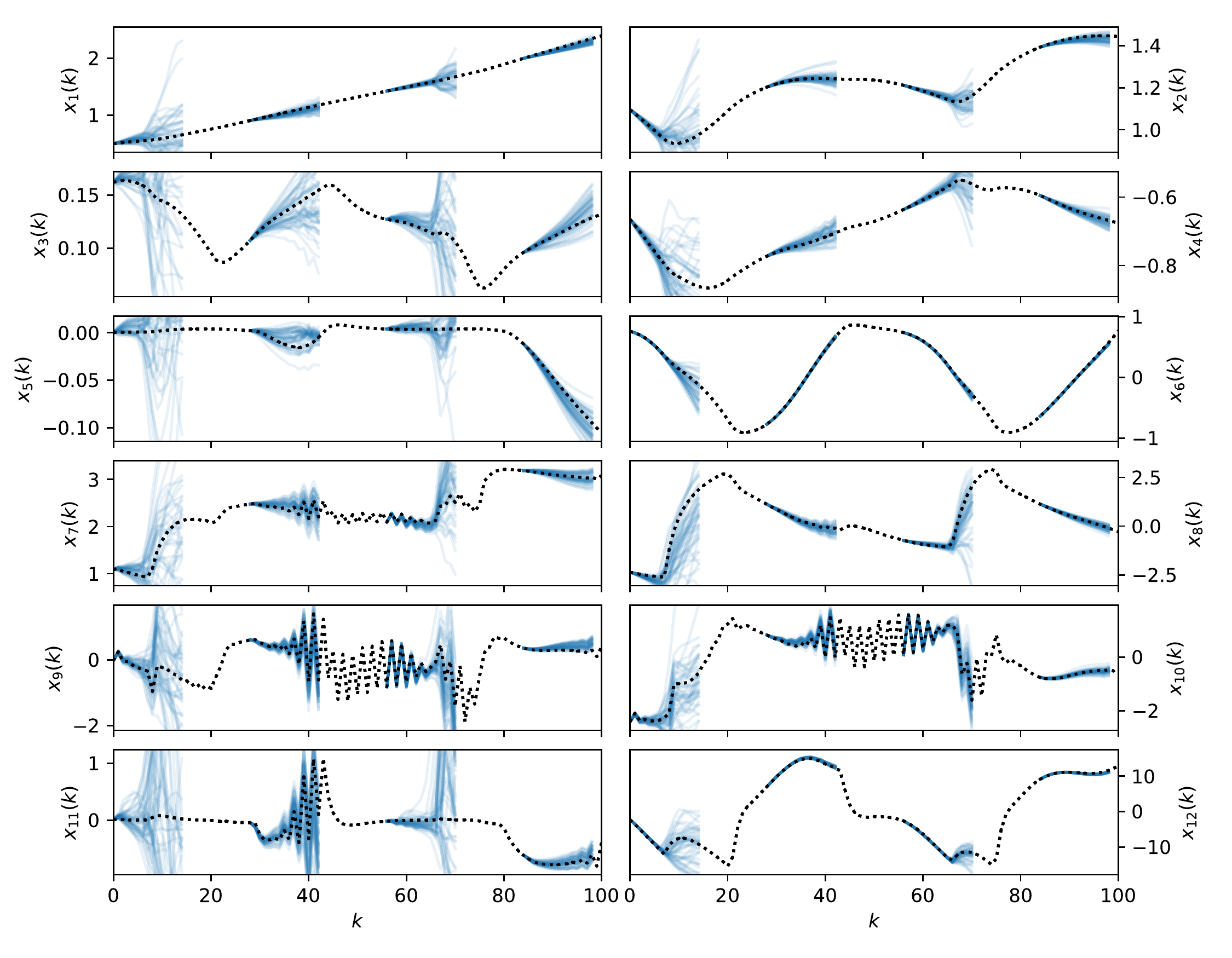}
    \caption{Rollouts for the hopper environment (dotted black) and predicted rollouts from dynamics models sampled from the posterior maintained by \algName{} (blue). As can be seen, as \algName{} incorporates online experience into its posterior, the uncertainty in the dynamics is reduced, with predicted trajectories clustering closer to the true trajectory.}
    \label{fig:hopper_rollout}
\end{figure}

For the hopper, a policy was trained via Proximal Policy Optimization \cite{schulman2017proximal} with the dynamics randomization described in the body of the paper. This was utilized to generate hopping behavior. Using this policy was necessary, as open loop simulation of the hopper results in the system just collapsing to the ground. Rollouts of \algName{} and MAML are plotted in figures \ref{fig:hopper_rollout} and \ref{fig:hopper_rollout_maml}. In this experiment, two hidden layers of $128$ units were used, with $32$ basis functions. For both the pendulum and the hopper, performance did not noticeably change with a larger number of basis functions. A comparison of time required to evaluate \algName{} versus GPR is plotted in figure \ref{fig:time}. Note that on the high dimensional hopper system, GPR rapidly becomes unsuitable for real time usage. While this performance could be improved with sparsification techniques, these are likely also unsuitable for use online. These timing curves were generated using a computer with a 3.6GHz octo-core AMD Ryzen 1800X CPU. 

\subsection{Vehicle Lane Change}

For this experiment, the lane change dataset from \cite{schmerling2017multimodal} was used. This dataset consists of 1105 episodes of humans pairs driving against each other (on a driving simulator), wherein the goal is to switch lanes with each other without colliding. The participants were not allowed to communicate. A total of 19 pairs of people are captured in the dataset. The data is captured at a frequency of 10Hz. While episodes are nominally 50 timesteps long, some end earlier than this. As a result, we cut all trajectories to the minimum length observed in the data, which was 33 timesteps. Again, two hidden layers of $128$ units each were used, and $32$ basis functions. Performance stayed relatively constant for a greater number of basis functions, but degraded for $16$. The noise $\Sigma_\epsilon$ was set as 
\begin{equation*}
\Sigma_\epsilon = \mathrm{diag}([0.001, 0.0005, 0.005, 0.0025, 0.001, 0.0005, 0.005, 0.005]).
\end{equation*}
These values were estimated by training the \algName{} with $\Sigma_\epsilon = 0.001 I$, calculating the MSE of each element on trajectories in a validation set when conditioned on the entire trajectory, and setting the diagonal elements of $\Sigma_\epsilon$ to be approximately these values.

The prediction task was, given the position and velocity of the vehicles, predict the next position and velocity of both vehicles. Predicting the position is relatively easy as the velocity may be used for first order approximations, but the velocity (specifically, the change in velocity) at the next timestep captures the actions taken by the drivers. As such, the task was effectively predicting human decision-making. We found performance was substantially improved by predicting the state difference between subsequent timesteps. The dataset was split into a train set of 1000 examples and a test set of 100. For details on the specific initial conditions of the dataset, as well as the mechanics of the simulator, we refer the reader to \cite{schmerling2017multimodal}. The initial conditions were chosen to make it ambiguous to the drivers whether passing in front of or behind the other vehicle is preferable, and so the dataset captures humans reasoning in real time under uncertainty. 

\begin{figure}

    \centering
    \includegraphics[width=0.49\columnwidth]{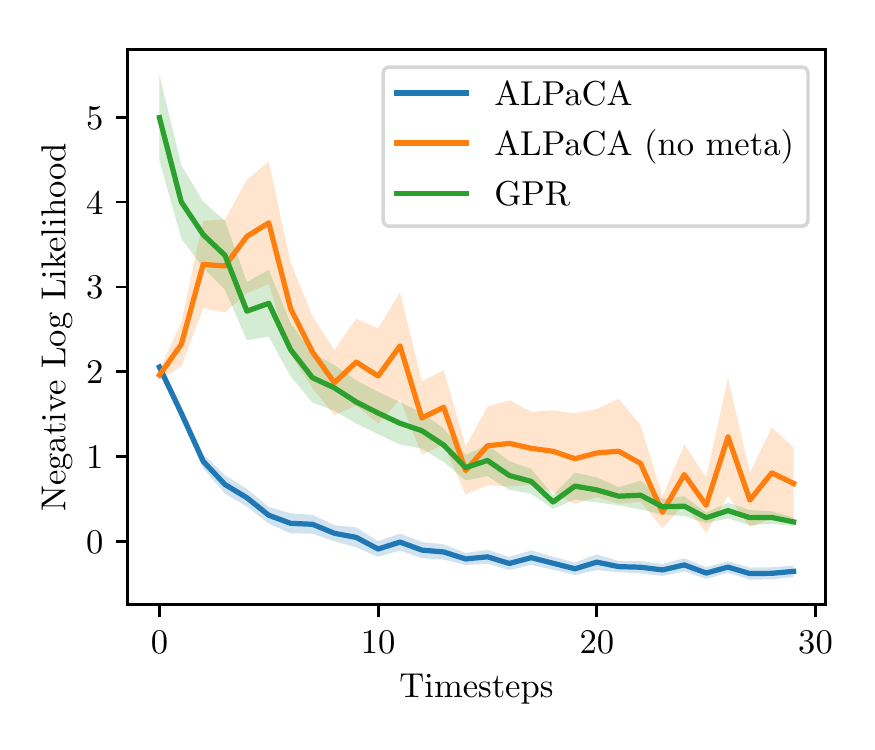}
    \includegraphics[width=0.49\columnwidth]{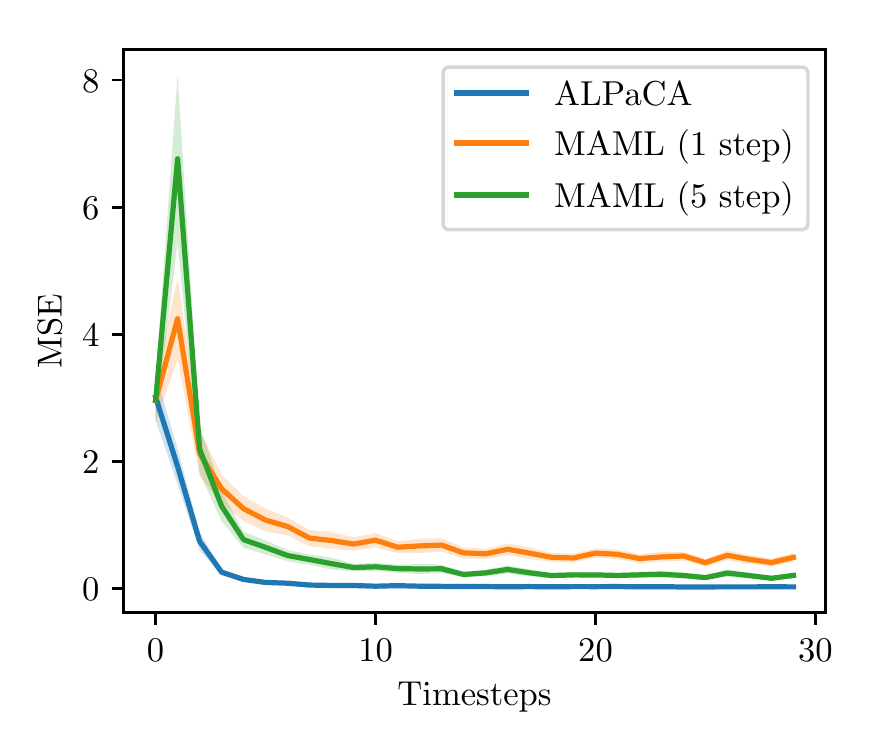}
    
    \caption{Negative Log Likelihood (left) and Mean Squared Error (right) for the sinusoid problem. }
    \label{fig:nll_toy1}
\end{figure}

\begin{figure}
    \centering
    \includegraphics[width=0.49\columnwidth]{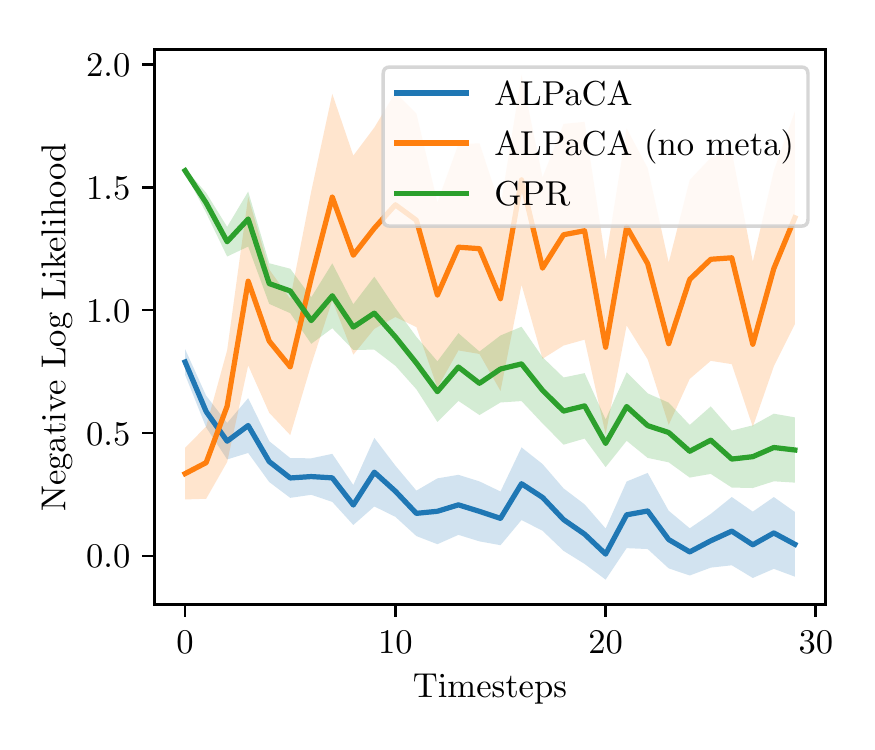}
    \includegraphics[width=0.49\columnwidth]{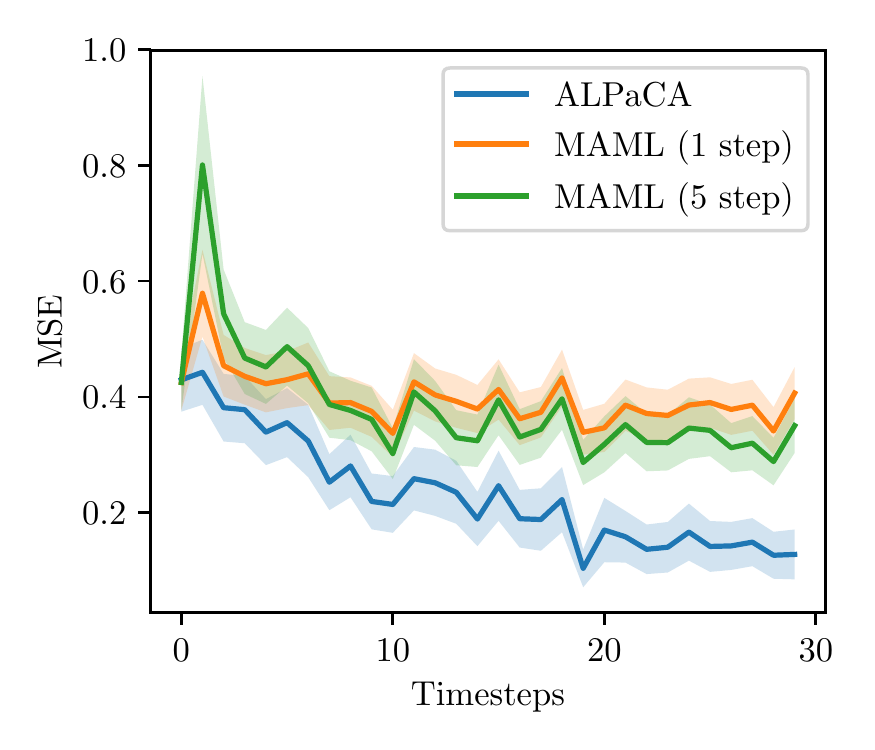}
    \caption{Negative Log Likelihood (left) and Mean Squared Error (right) for the step function problem.}
    \label{fig:nll_toy2}
\end{figure}

\begin{figure}
    \centering
    \includegraphics[width=\columnwidth]{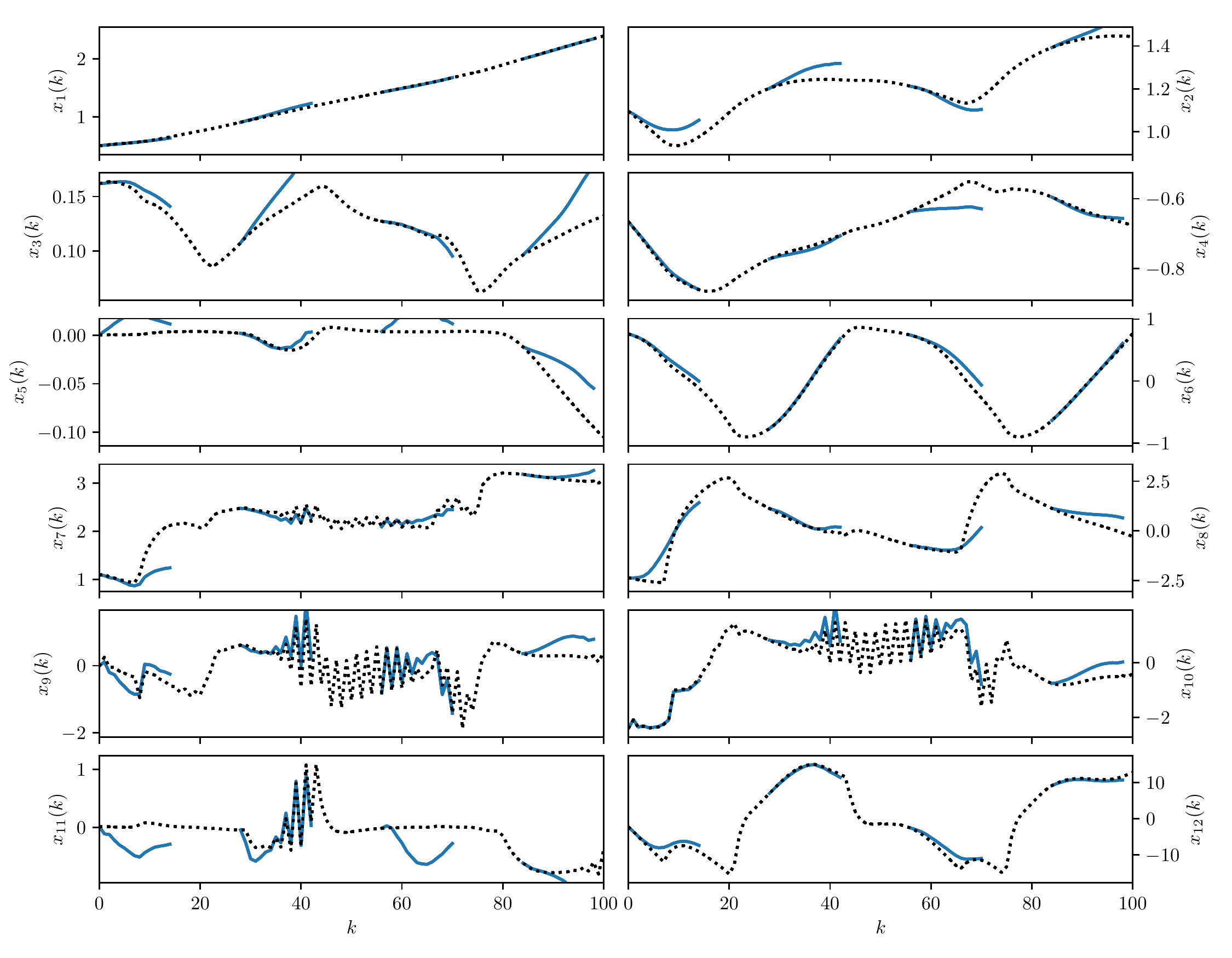}
    \caption{Rollouts for the hopper environment (dotted black) and predicted rollouts from MAML (blue).}
    \label{fig:hopper_rollout_maml}
\end{figure}

\begin{figure}
    \centering
    \includegraphics[width=\columnwidth]{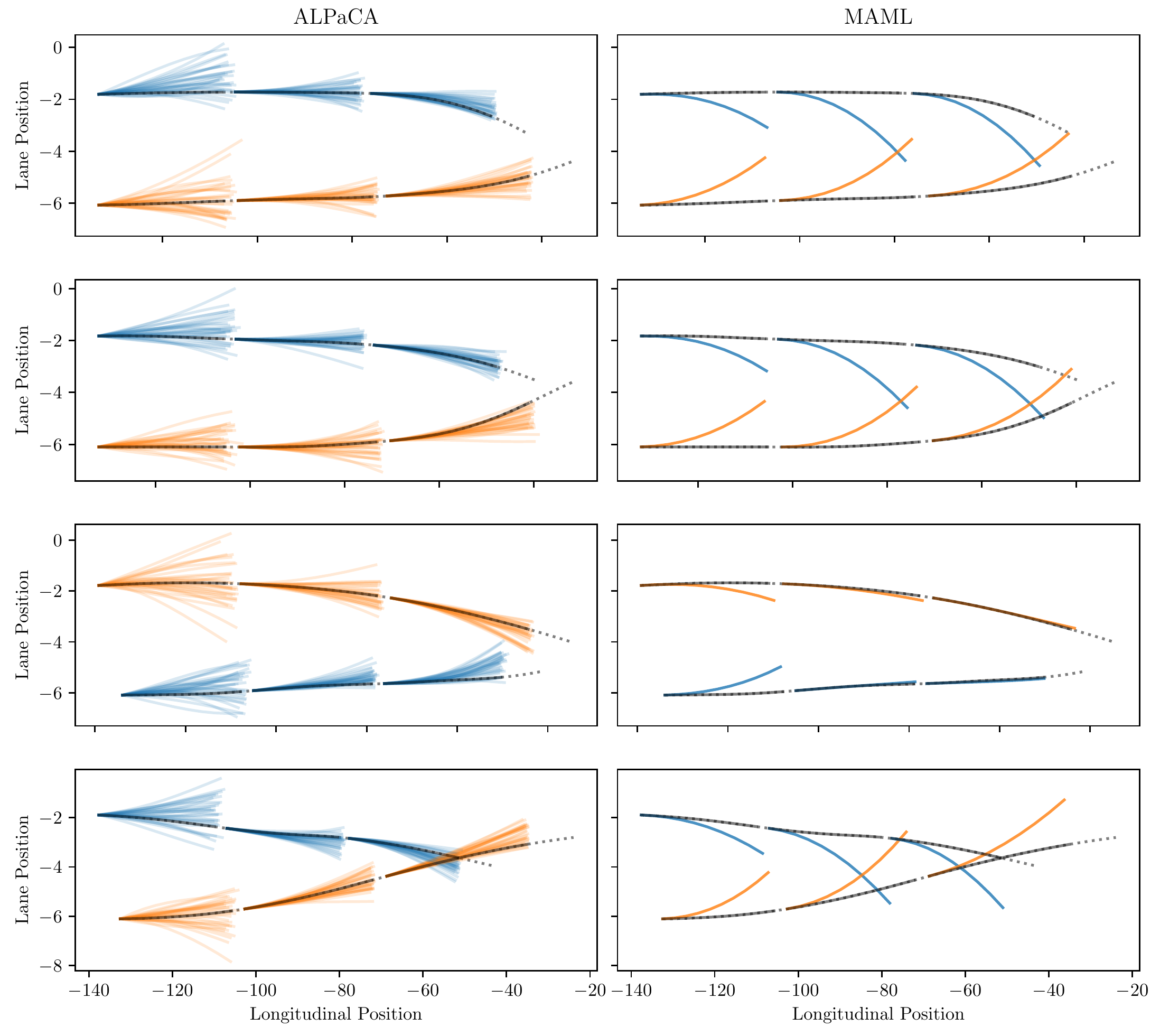}
    \caption{Rollouts for the lane-change example environment, each row corresponding to a different trial in the test set. Dotted black shows the whole trajectory of both cars. 10 timestep predictions by ALPaCA (left), and MAML (right) at timesteps 0, 10, and 20 are overlaid. These predictions are conditioned on the trajectory of both cars up to that point. The ground truth over the time period of each prediction is highlighted in solid black. \algName{} consistently reduces uncertainty and improves the quality of its predictions as more data is observed. While this happens in some cases with MAML, in many cases MAML fails to adapt, and furthermore, gives no indication of its uncertainty.}
    \label{fig:hopper_rollout_maml}
\end{figure}



\end{document}